\documentclass{article}
\PassOptionsToPackage{numbers,sort&compress}{natbib}
\usepackage[preprint]{neurips_2026}
\usepackage[T1]{fontenc}
\usepackage[utf8]{inputenc}
\usepackage{amsmath,amssymb,booktabs,graphicx,tabularx,longtable,multirow}
\usepackage{microtype,xcolor,url,needspace,wrapfig}
\usepackage{placeins}
\usepackage[breakable]{tcolorbox}
\usepackage{fvextra}
\usepackage{hyperref}

\hypersetup{colorlinks=true,linkcolor=blue!45!black,citecolor=blue!45!black,urlcolor=blue!45!black,
pdftitle={LOLPlay: Benchmarking Real-Time Visual Agents in League of Legends},pdfauthor={Zhaojian Yu}}
\newcommand{\benchmark}{LOLPlay}

\newcommand{\code}[1]{\texttt{#1}}

\title{LOLPlay: Benchmarking Real-Time Visual Agents\\in League of Legends}
\author{Zhaojian Yu \\ \texttt{zhaojian.yu01@gmail.com}}
\begin{document}
\maketitle

\begin{abstract}
Visual agents need to coordinate reasoning and action while their environment evolves during inference. We introduce \benchmark{}, a benchmark in native \emph{League of Legends} with 36 champion-specific tasks spanning resource management, combat, cooperation, and objectives. Agents interact through screenshots and mouse--keyboard actions; automated initialization and 82 executable subgoal checks distinguish partial progress from completion. Across 324 task evaluations of three models at three reasoning-effort settings, GPT-6 Astra achieves the highest task success rate of 25.0\% at low and medium effort, and the highest subgoal success rate of 40.3\% at medium effort. However, high effort improves neither metric over low effort for any model. For Astra, longer requests leave fewer interaction rounds, suggesting a cost to additional reasoning in real-time control. Trajectories reveal incomplete subgoal composition and actions executing after the observed scene changes, motivating agents that combine visual reasoning with timely action.\footnote{Project repository: \url{https://github.com/answers111/LOLPlay}.}
\end{abstract}
\begin{figure}[!ht]
\centering
\includegraphics[width=0.95\linewidth]{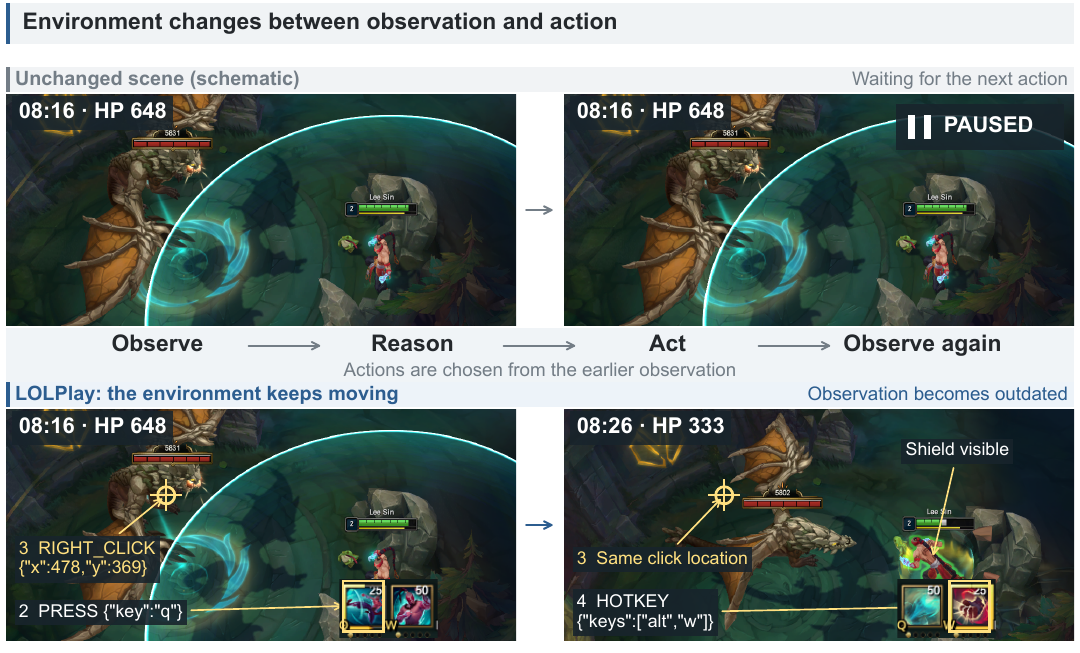}
\caption{\textbf{The agent thinks; the environment keeps moving.} Top: a schematic environment waiting for an action (e.g., VideoGameBench Lite~\cite{videogamebench}), not a paused-game experiment. Bottom: \benchmark{} advances approximately 10 game seconds across reasoning, actions, and capture.}
\label{fig:teaser}
\end{figure}

\clearpage
\begin{figure}[!ht]
\centering
\includegraphics[width=\linewidth]{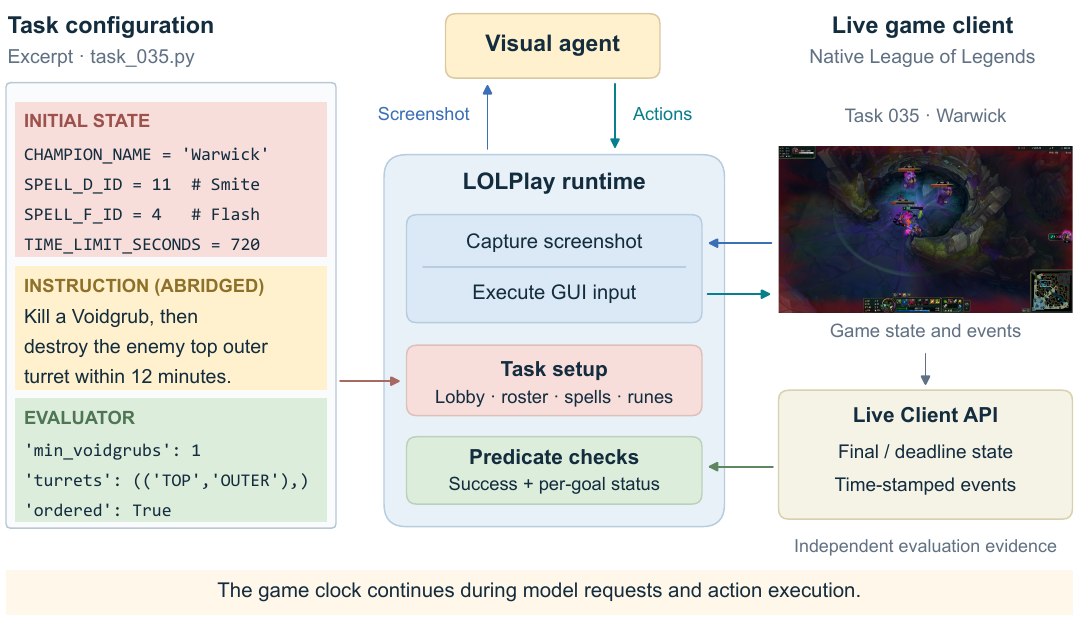}
\caption{\textbf{The \benchmark\ execution and evaluation workflow.} Task configuration connects screenshot-based interaction to independent state/event checks while gameplay continues. The abridged configuration and unmodified game frame illustrate task~035 (Warwick). Live Client API evidence supplies final/deadline states and time-stamped events for evaluation.}
\label{fig:overview}
\end{figure}
\section{Introduction}
\label{sec:intro}
An intelligent agent should be able to pursue a goal even when the world does not wait for its next decision. For visual agents, which use visual observations to guide their actions, this means reasoning about a scene that may change before they can act. Consider an agent preparing an action from a screenshot: by the time it finishes reasoning, the target may have moved, an opportunity may have disappeared, or a previously safe situation may have become dangerous. The challenge is to keep a sequence of decisions aligned with an environment that changes throughout execution. This raises a central question: \emph{can today's visual agents turn their understanding of a scene into timely actions that accomplish a sustained goal?}

To study this question, we introduce \benchmark{}, a benchmark for real-time visual agents in live \emph{League of Legends}. It evaluates whether agents can combine timely interaction with the completion of multiple task requirements while gameplay continues throughout reasoning and execution. Explicit task outcomes and subgoal checks make both complete success and the gap between local progress and unfinished goals measurable.

Progress in computer-use and game evaluation provides the foundation for this work. OSWorld~\cite{osworld} and OSWorld~2.0~\cite{osworld2} assess tasks in real applications and extended workflows; VideoGameBench~\cite{videogamebench} and GameWorld~\cite{gameworld} examine interaction with evolving game environments. \benchmark{} focuses on champion-specific objectives in a native multiplayer online battle arena (MOBA), where resource acquisition, combat, survival, and objective sequencing can constrain the same task. Table~\ref{tab:related} compares task settings and evaluation evidence.

\begin{table}[!t]
\centering
\small
\setlength{\tabcolsep}{3.5pt}
\caption{\textbf{Comparison with computer-use and game benchmarks.} \benchmark\ combines champion-specific tasks in a native MOBA with uninterrupted gameplay and state- and event-based evaluation. Real-time interaction and executable scoring individually have precedents; the table compares task settings and evaluation evidence, not model scores.}
\label{tab:related}
\begin{tabularx}{\linewidth}{@{}>{\raggedright\arraybackslash}p{1.23in}*{2}{>{\raggedright\arraybackslash}X}@{}}
\toprule
\textbf{Benchmark} & \textbf{Environment}  & \textbf{Evaluation} \\
\midrule
OSWorld~\cite{osworld} & Desktop; 369 tasks & Execution-based checks \\
OSWorld 2.0~\cite{osworld2} & Desktop/web; 108 workflows & Functional checkpoints; bounded judging \\
VideoGameBench~\cite{videogamebench} & 10 test games & Game progress milestones \\
GameWorld~\cite{gameworld} & Browser; 34 games, 170 tasks & Serialized state; success and progress \\
GameVerse~\cite{gameverse} & 15 games & VLM milestone evaluation \\
\textbf{LOLPlay} & Native LOL, unpaused; 36 tasks & Live Client state; event deadlines \\
\bottomrule
\end{tabularx}
\end{table}

The MOBA setting makes these demands concrete. A successful action is rarely an isolated achievement: reaching a target, choosing an ability, and preserving enough health to continue must work together. Meanwhile, units move and combat proceeds without waiting for the agent. An action planned for one encounter may arrive after that encounter has ended (Figure~\ref{fig:teaser}). Even when an agent secures an intermediate objective, it must adapt its subsequent decisions to complete the remaining task. These episodes expose whether the agent can adjust its behavior as the conditions for success change.

To assess these requirements beyond a single match-win outcome, \benchmark{} defines 36 champion-specific tasks across farming, development, survival, combat, cooperation, and map objectives. These tasks turn broad gameplay competence into explicit goals with identifiable requirements. The agent acts through the game's visual interface while gameplay continues during reasoning and execution, so the consequences of delay remain part of the interaction.

The benchmark connects this continuous interaction to task-level and subgoal-level evaluation. Each task has a defined starting configuration and executable success conditions; 82 subgoal checks identify which requirements are satisfied in each evaluated run. Independent game-state and event evidence records resource and skill requirements as well as the participants, timing, and order of objectives. This makes it possible to study both whether an agent succeeds and where partial progress stops short of the goal. Figure~\ref{fig:overview} summarizes the setup, interaction, and evaluation workflow.

Evaluation of three models at three reasoning-effort settings yields 324 task runs. GPT-6 Astra at low and medium effort shares the highest task success rate of 25.0\%; medium effort achieves the highest subgoal success rate of 40.3\%. Of the 40 successful runs, 38 concern economy or development and two satisfy a combat-category task's level and survival requirements. No run meets a personal-kill or allied-assist check, and 108 failed runs nevertheless satisfy a proper subset of requirements. Increasing reasoning effort does not consistently improve task completion: Astra ties at low and medium effort before declining at high effort, while the other two models perform best at low effort. Trajectory analysis further illustrates actions executing after the observed scene changes. Together, these findings expose the difficulty of maintaining effective behavior across an evolving task, beyond making progress on individual requirements.

Our contributions are \textbf{(1)} a benchmark of 36 tasks that brings sustained, multi-skill visual control into an uninterrupted native MOBA; \textbf{(2)} executable task and subgoal evaluation that makes both complete outcomes and partial progress measurable; and \textbf{(3)} an empirical study connecting task performance to reasoning configurations, execution budgets, and observation--action mismatch. \benchmark{} provides a testbed for developing agents whose reasoning remains useful as the world changes.

\section{LOLPlay Benchmark and Evaluation}
\label{sec:benchmark}
\subsection{Task Definition}
A \emph{task} asks an agent to accomplish a specified objective by controlling a champion, the player-controlled character, through the game interface. As in computer use, the agent observes screenshots and operates a mouse and keyboard. Right-clicking a ground location moves the champion; right-clicking an enemy requests an attack; pressing an ability key and selecting a target requests a spell; opening the shop and selecting an item purchases equipment when the game permits it. These familiar inputs must be coordinated while the game continues to evolve.

We specify task $i$ as
\begin{equation}
\tau_i=(I_i,C_i,B_i,\mathcal{G}_i,E_i),\qquad
\mathcal{G}_i=\{g_{i1},\ldots,g_{im_i}\},
\label{eq:task}
\end{equation}
where $I_i$ is the instruction, $C_i$ is the starting configuration, $B_i$ is the time budget, $\mathcal{G}_i$ is the set of required subgoals, and $E_i$ is the task evaluator. A \emph{subgoal} is one individually checked requirement, such as reaching a level, learning an ability, or destroying a target. Task success requires satisfying all executable requirements, including applicable event-time and ordering checks, under the sampling protocol in Section~\ref{sec:protocol}. For example, the Cho'Gath task has three subgoals: reach level~6, learn the ultimate ability, and increase maximum health by at least 350. One attempt at a task is an \emph{episode}.

The suite contains 36 tasks, one per champion, with 82 subgoals and budgets of 5--20 minutes. Economy tasks concern resource acquisition; development tasks concern ability allocation and attribute growth; sustain tasks concern maintaining health, mana, or survival. Combat tasks involve fighting or surviving against enemy champions, cooperation tasks involve allied-bot kills, and map-objective tasks target structures or neutral monsters. Table~\ref{tab:families} summarizes these six families; their requirements can overlap.

\begin{table}[!htbp]
\centering
\small
\setlength{\tabcolsep}{4pt}
\caption{\textbf{Current task suite.} Groups describe primary demands; requirements overlap. Subgoal counts and all reported results use the current task definitions. The complete catalog appears in Appendix~\ref{app:catalog}.}
\label{tab:families}
\begin{tabularx}{\linewidth}{@{}lrrX@{}}
\toprule
\textbf{Primary task demand} & \textbf{Tasks} & \textbf{Subgoals} & \textbf{Examples of outcomes} \\
\midrule
Economy and farming & 6 & 7 & Last hits, level and gold, purchases \\
Development  & 9 & 26 & Ability allocation, attribute growth \\
Sustain & 7 & 18 & Health, mana, survival to a deadline \\
Combat & 6 & 13 & Kills, level, survival constraints \\
Cooperation & 3 & 8 & Assists and participation with bots \\
Map objectives & 5 & 10 & Dragon, turret, ordered objectives \\
\midrule
Total & 36 & 82 & 36 distinct champions; 5--20 minute budgets \\
\bottomrule
\end{tabularx}
\end{table}

\subsection{Agent Interface}
\begin{wraptable}{r}{0.60\textwidth}
\vspace{-8pt}
\centering
\footnotesize
\setlength{\tabcolsep}{4pt}
\renewcommand{\arraystretch}{1.0}
\caption{Example actions in function notation, following the presentation in OSWorld~\cite{osworld}. Structured equivalents and parameters appear in Appendix~\ref{app:action-termination}.}
\label{tab:action-space}
\begin{tabularx}{\linewidth}{@{}>{\raggedright\arraybackslash}p{.43\linewidth}>{\raggedright\arraybackslash}X@{}}
\toprule
\textbf{Function} & \textbf{Description} \\
\midrule
\code{moveTo(x, y)} & Move the pointer to $(x,y)$. \\
\code{click(x, y)} & Click at $(x,y)$. \\
\code{typewrite('text')} & Type in the focused field. \\
\code{press('q')} & Press the Q key. \\
\code{hotkey('ctrl', 'q')} & Press Ctrl and Q together. \\
\code{scroll(200)} & Scroll up by 200 units. \\
\code{scroll(-200)} & Scroll down by 200 units. \\
\code{dragTo(x, y)} & Drag the pointer to $(x,y)$. \\
\code{keyDown('shift')} & Hold the Shift key down. \\
\code{keyUp('shift')} & Release the Shift key. \\
\code{WAIT} & Wait; the game stays live. \\
\code{FAIL} & Declare inability to proceed. \\
\code{DONE} & Declare task completion. \\
\bottomrule
\end{tabularx}
\vspace{-6pt}
\end{wraptable}

At decision time $t_k$, the agent receives a screenshot $o_k$, instruction $I_i$, and recent observation--action history $h_k$, and returns one or more mouse--keyboard actions $a_k\sim\pi(\cdot\mid I_i,h_k,o_k)$. Table~\ref{tab:action-space} lists examples. The observation does not include a separate list of units and their coordinates or health, or the game-state data used by the evaluator. The agent should read relevant information from the screenshot before acting.

The environment keeps running while the agent generates its response and executes the actions. The scene at execution time can therefore differ from $o_k$: an enemy may have moved, combat may have ended, or a dialog may obstruct the intended click (Figure~\ref{fig:teaser}). This observation--action delay is part of the task rather than a pause in the game.
Table~\ref{tab:task-examples} pairs each task family with an observation, an instruction, and a simplified evaluation script.
\begin{table}[!htbp]
\centering
\setlength{\tabcolsep}{4pt}
\renewcommand{\arraystretch}{1.12}
\small
\caption{\textbf{Representative tasks and evaluation scripts.} Screenshots illustrate interaction, not necessarily success. Pseudocode abbreviates data retrieval and validation; event helpers select the specified actors and targets within the elapsed game-time limit (seconds). \texttt{deadline\_state} denotes the execution-deadline snapshot.}\label{tab:task-examples}
\medskip
\begin{tabular}{@{}>{\raggedright\arraybackslash}m{.30\linewidth}>{\raggedright\arraybackslash}m{.30\linewidth}>{\raggedright\arraybackslash}m{.36\linewidth}@{}}
\toprule
\textbf{Observation} & \textbf{Task Instruction (Abridged)} & \textbf{Evaluation Script (Simplified)}\\
\midrule
\textbf{Economy: Syndra (036)}\par\smallskip\includegraphics[width=\linewidth]{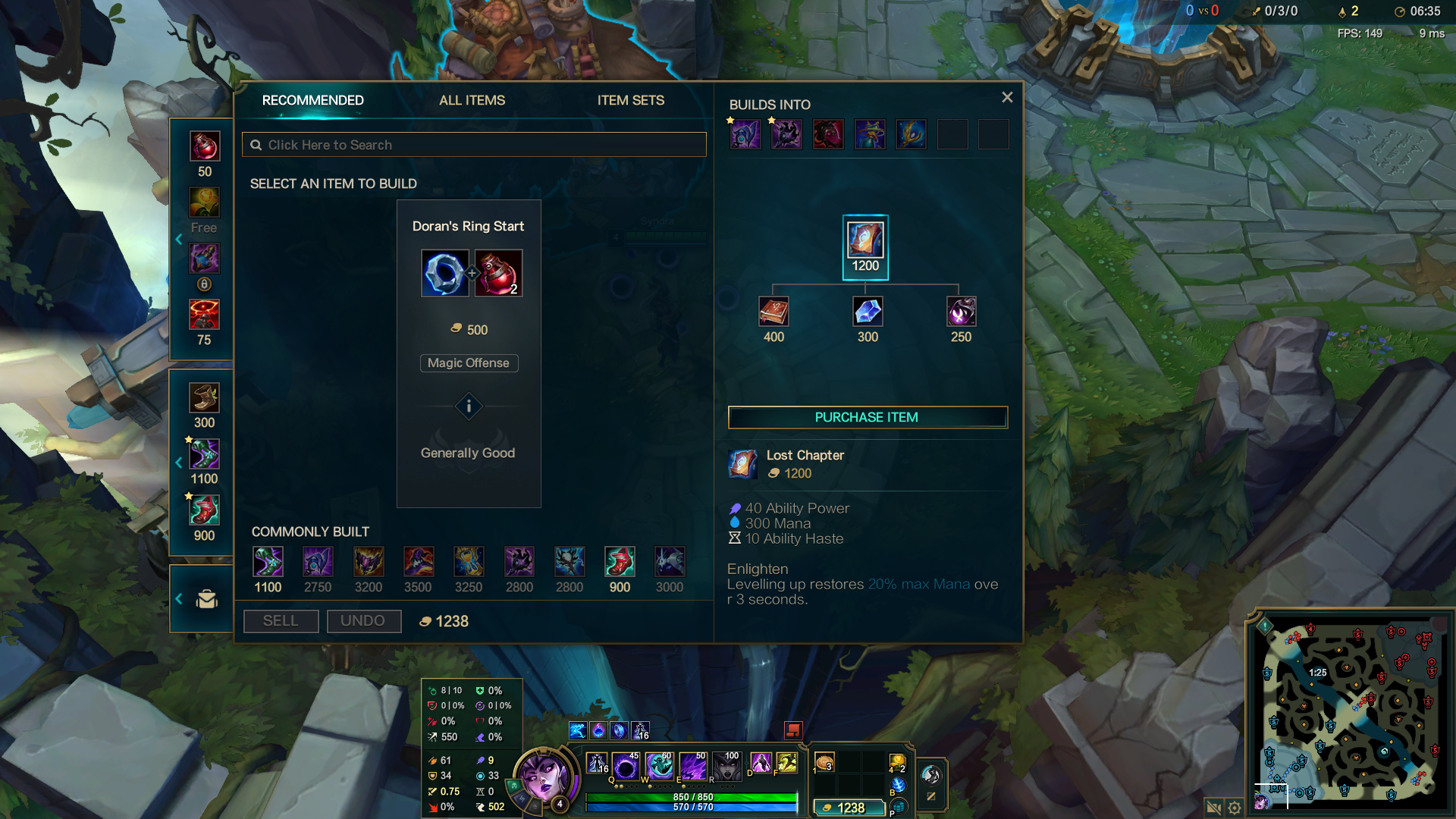} & \textit{Buy Lost Chapter within 7 minutes; finish while it is in the inventory.} & \begin{minipage}{\linewidth}\footnotesize\ttfamily player = live\_state()\par lost\_chapter = 3802\par assert lost\_chapter in (\par \hspace*{1em}player.item\_ids)\end{minipage} \\[5pt]
\midrule
\textbf{Development: Ahri (009)}\par\smallskip\includegraphics[width=\linewidth]{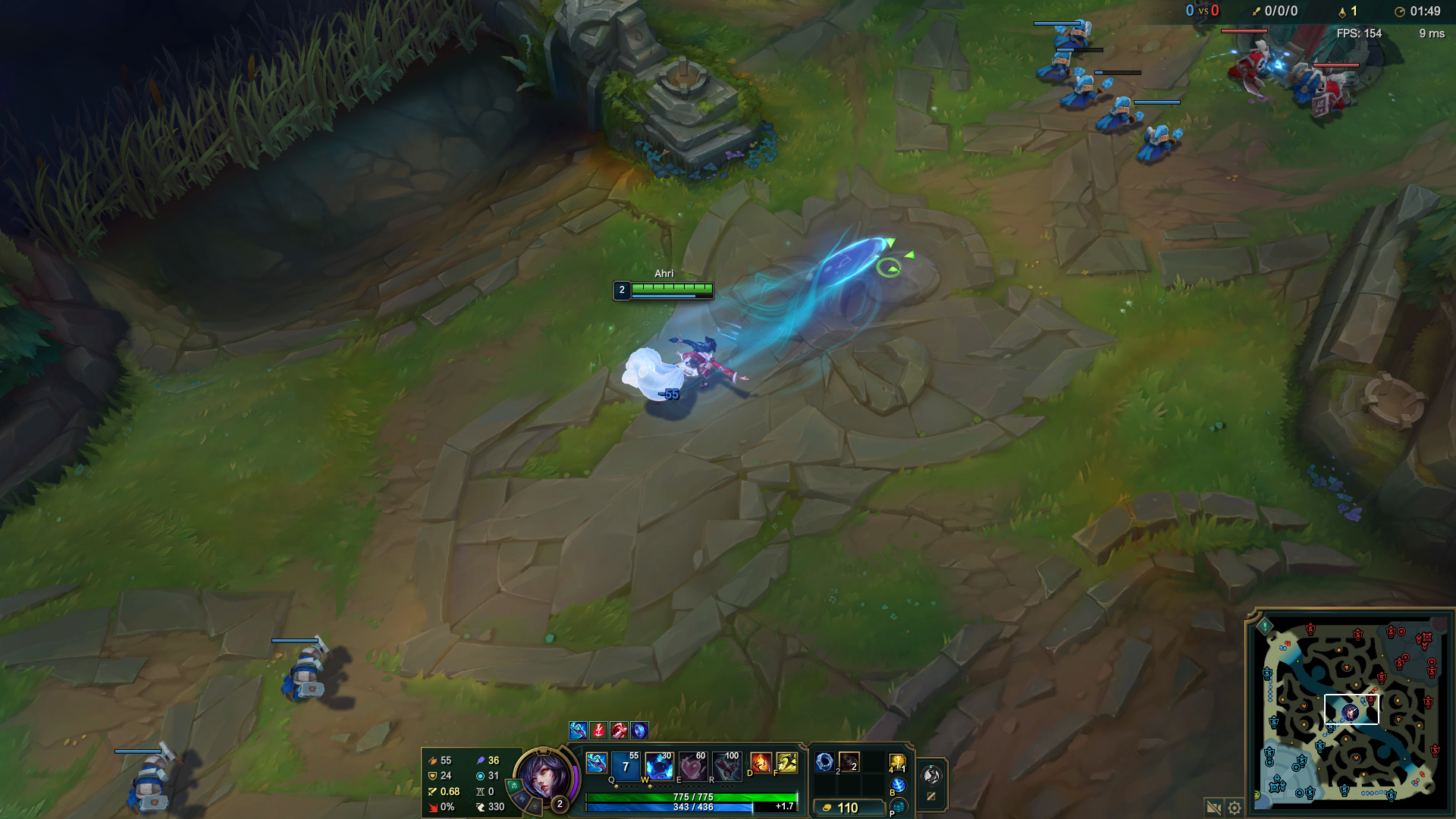} & \textit{Reach 10 CS and level 4 within 7 minutes, finishing with Q2, W1, E1 and R0.} & \begin{minipage}{\linewidth}\footnotesize\ttfamily player = live\_state()\par assert player.cs >= 10\par assert player.level >= 4\par assert player.qwer == (2,1,1,0)\end{minipage} \\[5pt]
\midrule
\textbf{Sustain: Brand (016)}\par\smallskip\includegraphics[width=\linewidth]{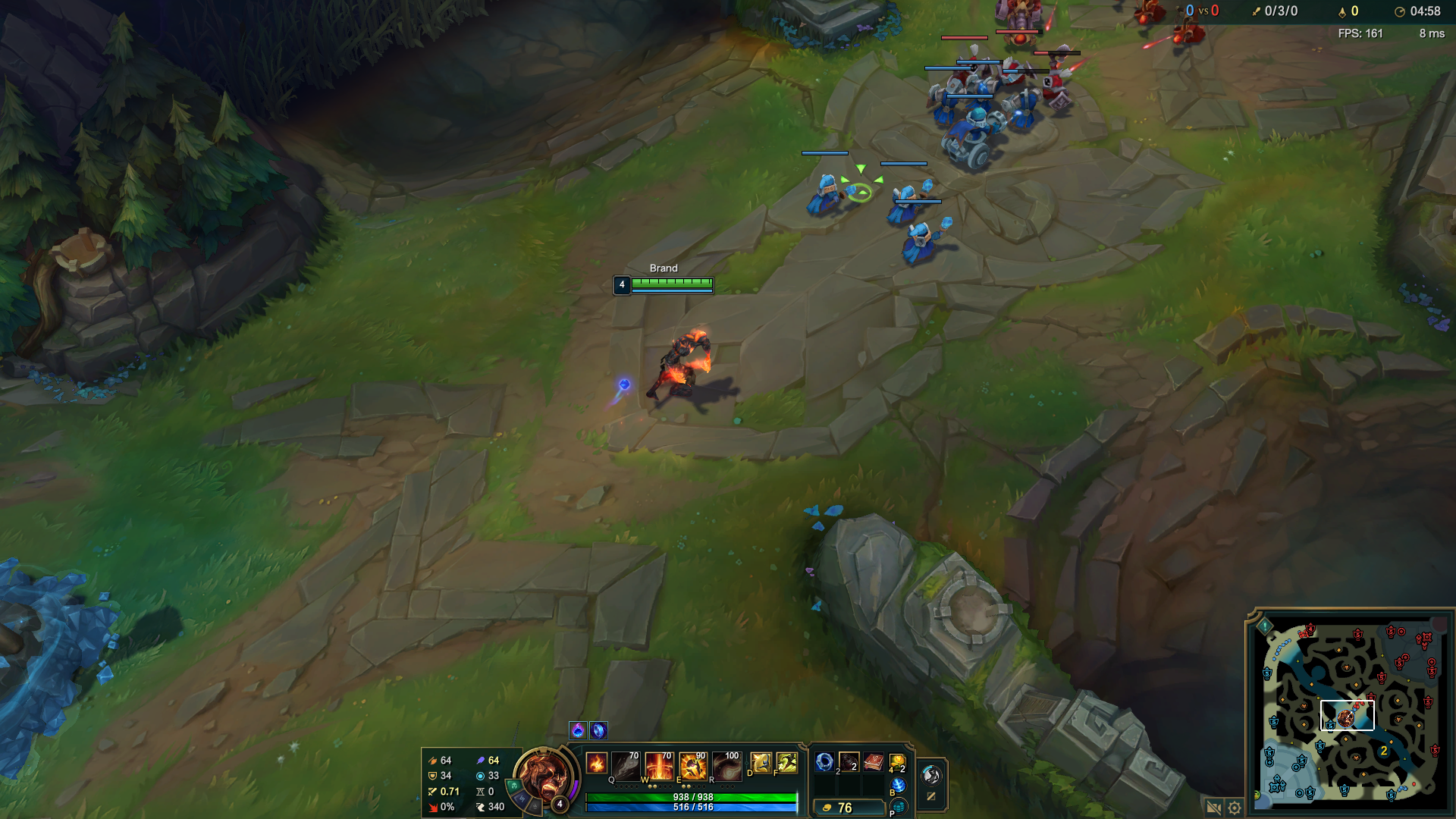} & \textit{At the end of 8 minutes, have at least 20 CS and 50\% mana. Keep playing until the deadline.} & \begin{minipage}{\linewidth}\footnotesize\ttfamily player = deadline\_state(480)\par assert player.cs >= 20\par assert player.mana\_ratio >= 0.5\end{minipage} \\[5pt]
\midrule
\textbf{Combat: Darius (022)}\par\smallskip\includegraphics[width=\linewidth]{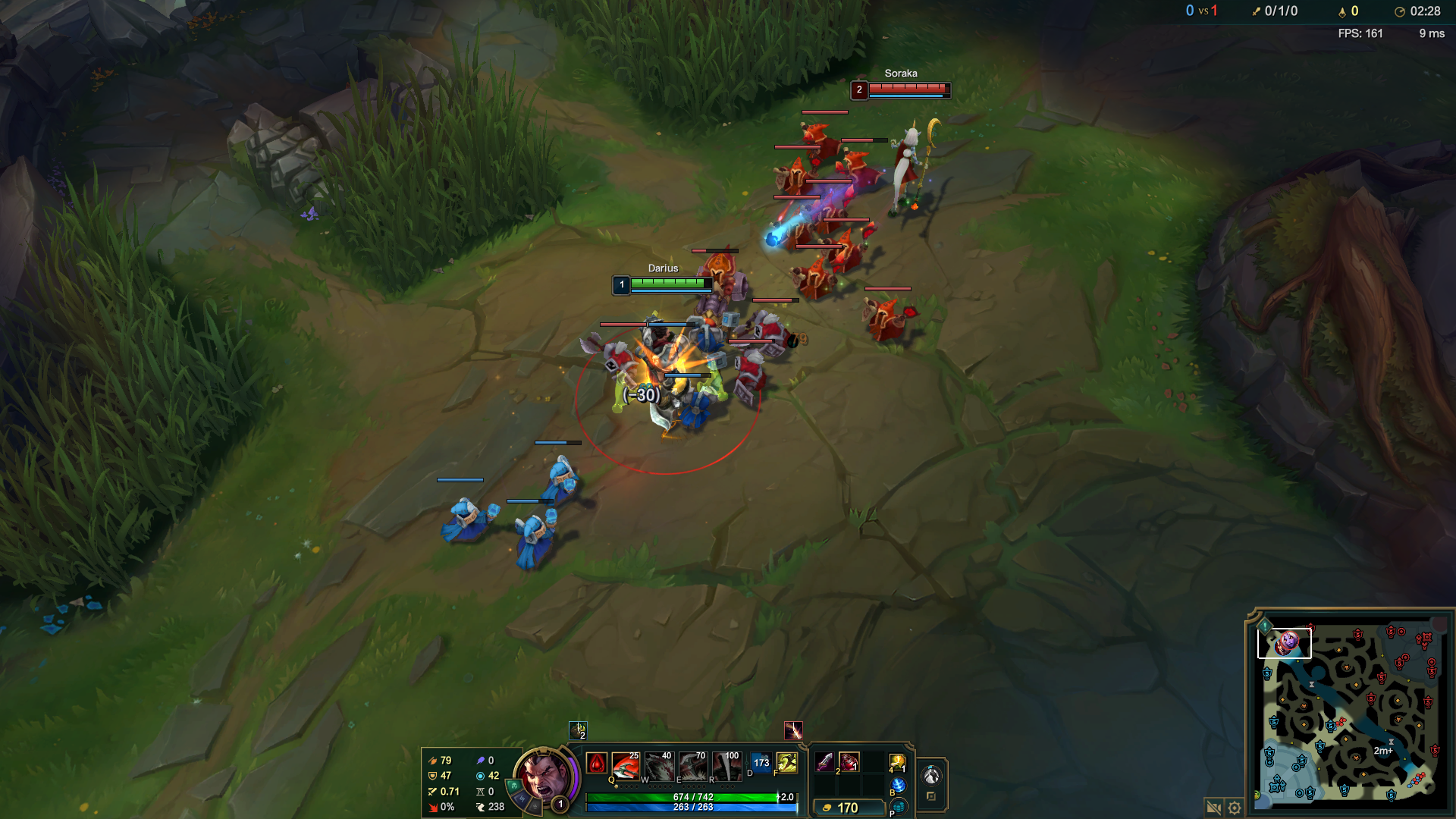} & \textit{Personally kill the enemy Soraka bot at least once within 9 minutes, with no more than one death.} & \begin{minipage}{\linewidth}\footnotesize\ttfamily player = live\_state()\par kills = personal\_kills(540)\par assert len(kills) >= 1\par assert player.deaths <= 1\end{minipage} \\[5pt]
\midrule
\textbf{Cooperation: Lulu (025)}\par\smallskip\includegraphics[width=\linewidth]{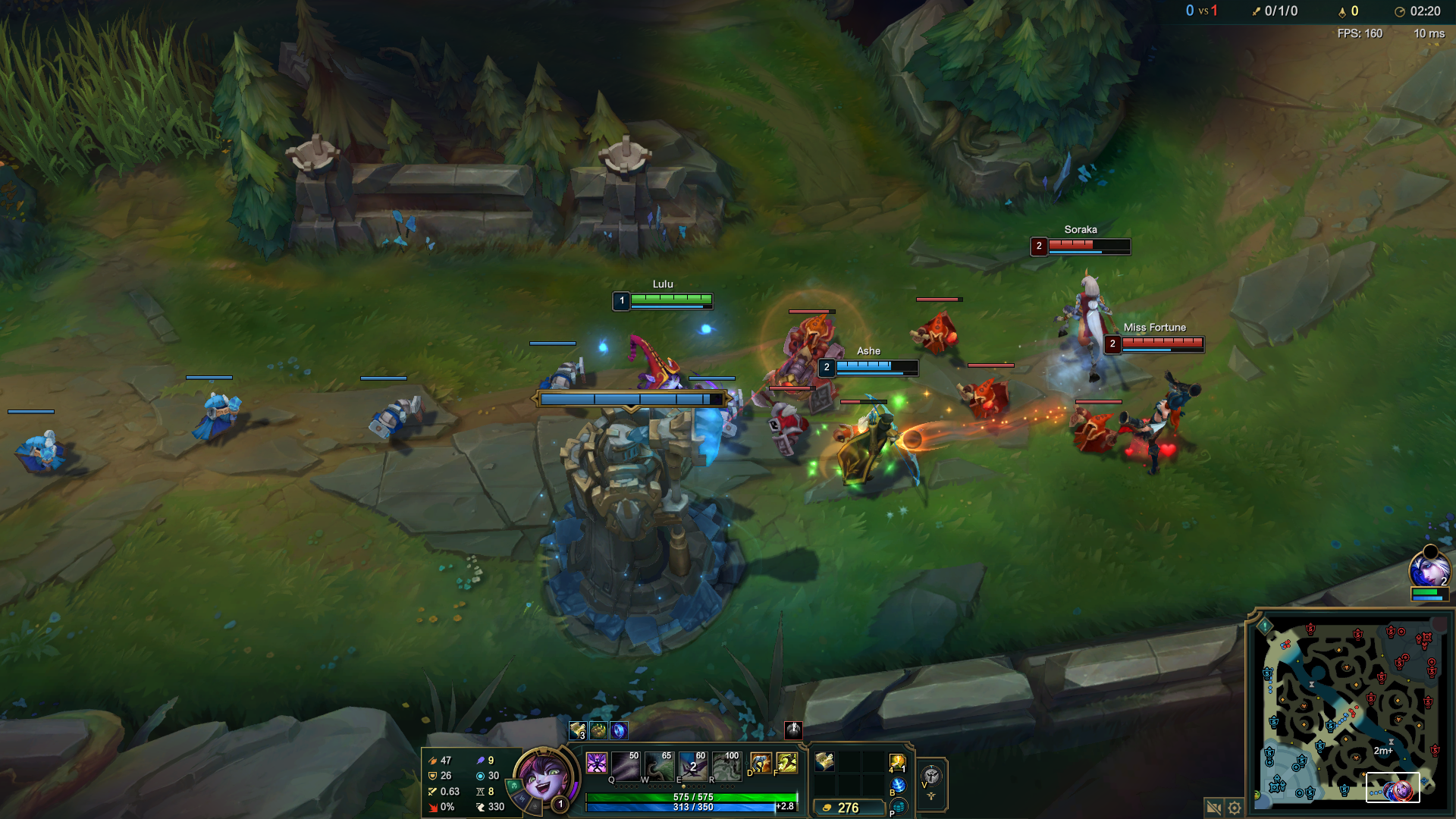} & \textit{Help allied Ashe secure a kill against Miss Fortune or Soraka within 10 minutes, with at most one death.} & \begin{minipage}{\linewidth}\footnotesize\ttfamily player = live\_state()\par kills = allied\_kills(600)\par assert any(\par \hspace*{1em}player.name in kill.assisters\par \hspace*{1em}for kill in kills)\par assert player.deaths <= 1\end{minipage} \\[5pt]
\midrule
\textbf{Map objectives: Warwick (035)}\par\smallskip\includegraphics[width=\linewidth]{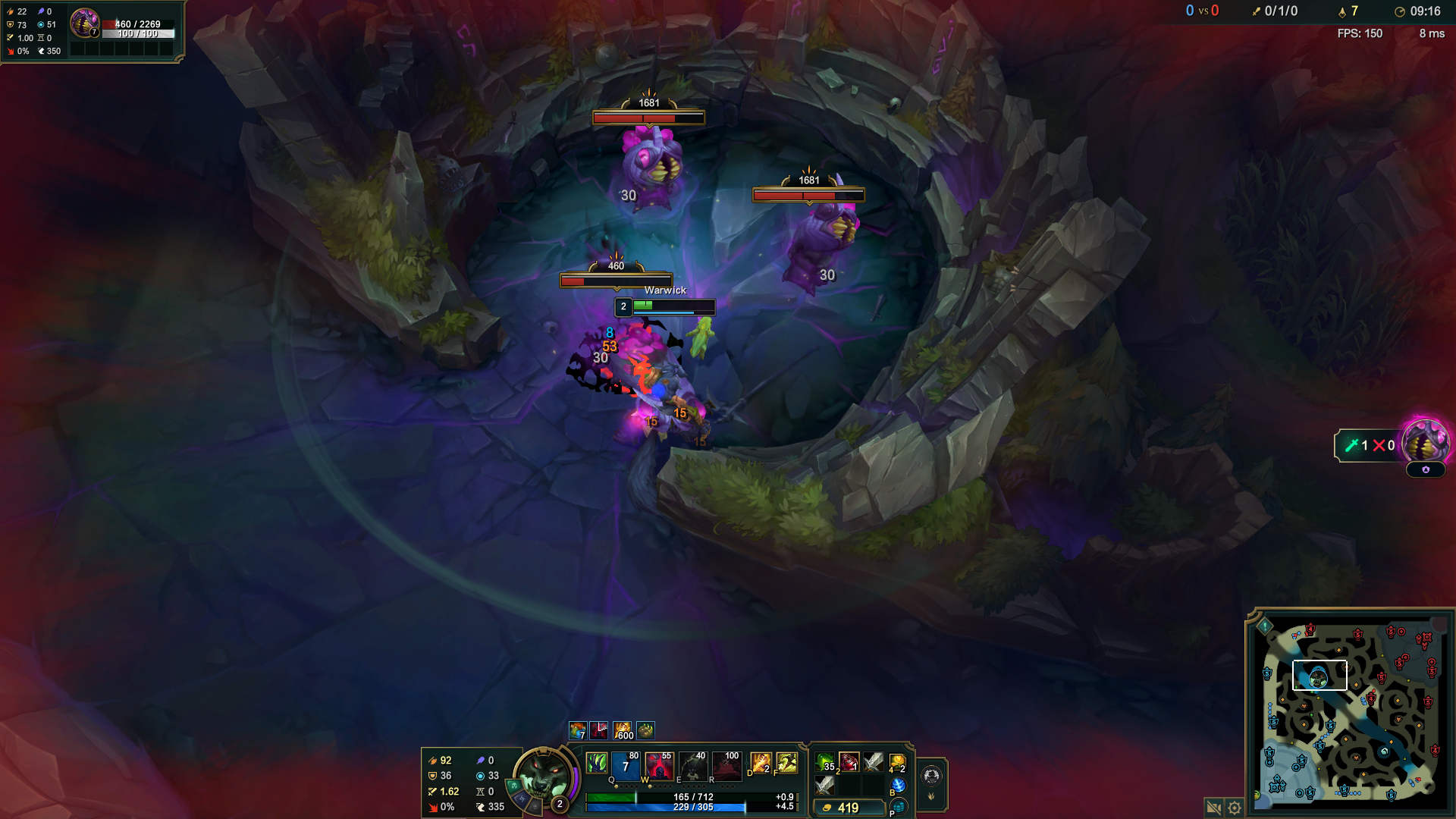} & \textit{Personally kill a Voidgrub, then destroy the enemy top outer turret within 12 minutes.} & \begin{minipage}{\linewidth}\footnotesize\ttfamily grubs = personal\_grub\_kills(720)\par turret = enemy\_top\_outer\_kill(\par \hspace*{1em}720)\par assert grubs and turret\par assert (min(grubs.times)\par \hspace*{1em}< turret.time)\end{minipage} \\[5pt]
\bottomrule
\end{tabular}
\end{table}
 Twenty-six tasks of LOLPlay contain no bot champions, but minions, turrets, monsters, resources, and the game clock remain active. Ten tasks include enemy bots; three also include an allied bot. Bots use beginner difficulty. Cooperation means participating in allied-bot kills, not communicating with another learned agent. These settings test basic control and multi-step objectives before full-match competition.

\subsection{Environment Setup}
We use the local League Client Update (LCU) API to prepare a fresh match from configuration $C_i$.\footnote{Riot Games: \href{https://developer.riotgames.com/docs/lol\#league-client-api}{League Client API documentation}.} Preparation finishes before the agent begins interacting with the game. Figure~\ref{fig:overview} shows the workflow; Table~\ref{tab:api-interfaces} summarizes the setup operations and the evaluation operations described below.

\begin{table}[!htbp]
\centering
\small
\setlength{\tabcolsep}{4pt}
\renewcommand{\arraystretch}{1.15}
\caption{\textbf{APIs used for setup and evaluation.} Each row pairs an API family with its usage. LCU configures the client; Live Client Data (LCD) supplies in-game evidence. Paths denote family prefixes for LCU and the queried endpoint for LCD. Neither service's payload is part of the standard agent observation. Operations and checker fields are listed in Appendix~\ref{app:lcu-interfaces}.}
\label{tab:api-interfaces}
\begin{tabularx}{\linewidth}{@{}>{\raggedright\arraybackslash}p{.12\linewidth}>{\raggedright\arraybackslash}p{.38\linewidth}>{\raggedright\arraybackslash}X@{}}
\toprule
\textbf{Stage} &\textbf{API family} & \textbf{Usage in LOLPlay} \\
\midrule
Setup & Lobby and bots (LCU)\newline \code{/lol-lobby/} & Create the custom lobby; set map and mode; add and verify bots; start champion selection. \\
\addlinespace[3pt]
Setup & Champion selection (LCU)\newline \code{/lol-champ-select/v1/} & Select and lock the champion; configure summoner spells and the default skin. \\
\addlinespace[3pt]
Setup & Rune pages (LCU)\newline \code{/lol-perks/v1/} & Update and activate the prescribed rune page. \\
\addlinespace[3pt]
Setup & Input settings (LCU)\newline \code{/lol-game-settings/v1/} & Set Q/W/E/R to normal casting; save and verify these settings. \\
\addlinespace[3pt]
Setup & Game phase (LCU)\newline \code{/lol-gameflow/v1/} & Poll the game phase until \code{InProgress}. \\
\midrule
Setup & Live game evidence (LCD)\newline \code{/liveclientdata/allgamedata} & Validate champion, roster, map, and initial conditions; retain required attribute baselines. \\
\addlinespace[3pt]
Evaluator & Live game evidence (LCD)\newline \code{/liveclientdata/allgamedata} & Read state snapshots and timestamped events for final-state, deadline, attribution, and ordering checks. \\
\bottomrule
\end{tabularx}
\end{table}

\paragraph{Client configuration through LCU.}
The setup routine connects to the authenticated local League client and creates a custom Summoner's Rift lobby (map 11, \code{CLASSIC}). It inserts the required allied and enemy bots, verifies their champions and teams, and starts champion selection. It then selects and locks the controlled champion, configures the two summoner spells and rune page, selects the default skin, and sets Q/W/E/R to normal casting. These are preparation operations; LCU is not the agent's combat-action interface. Lane preferences do not move the champion automatically.

\paragraph{Live-state validation and episode start.}
Setup waits for the client to enter \code{InProgress} and for the Live Client Data service to become available. It validates the champion, map, mode, roster, and task-specific initial conditions, such as level, inventory, creep score, deaths, or unlearned abilities. Tasks that measure growth also retain an initial attribute baseline. Only then does the runner record the task start and begin the execution budget; preparation time is excluded. Each task attempt starts in a newly launched match with the prescribed initial settings. Appendix~\ref{app:lcu-interfaces} lists the API operations for reproducibility.

\subsection{Evaluation}
The task evaluator $E_i$ checks whether the required outcomes occurred using the Live Client Data API, which provides player attributes, game time, and recorded events.\footnote{Riot Games: \href{https://developer.riotgames.com/docs/lol\#game-client-api_live-client-data-api}{Live Client Data API documentation}.} For example, it checks the player's level and learned abilities for development tasks, or identifies the killer and event time for a combat objective. An agent's completion message does not determine success. The API data is used for evaluation and is not included in the agent's screenshot observation.

\paragraph{Checking the task requirements.}
Each task has a dedicated evaluation function. Table~\ref{tab:task-examples} illustrates how live-state and event data become success checks. Syndra's evaluator checks the final inventory, while Ahri's jointly checks creep score, level, and exact ability ranks. Brand's evaluator uses the deadline snapshot to check both creep score and remaining mana. Combat and cooperation checks additionally identify who killed or assisted whom; Warwick's evaluator verifies that a personal Voidgrub kill precedes destruction of the enemy top outer turret, with both events inside the time limit. These checks distinguish reaching a state from completing a sequence of events. Appendix~\ref{app:lcu-interfaces} describes the supporting API data.

The evaluator returns a binary task result $Y_i$ and a result for each subgoal. A task is successful only when its full objective is verified, including any required timing and event order. Reporting subgoals separately shows which requirements were achieved when the task was not completed. Section~\ref{sec:experiments} defines how these results are aggregated into task success rate and subgoal success rate.

\paragraph{Human feasibility validation.}
Two players, each with more than 200 previously played \emph{League of Legends} matches, validate all 36 tasks using the same task configurations, time budgets, and executable evaluator as the agents. All 36 tasks were successfully completed (100\% task completion). This task-level validation establishes that the objectives are achievable by experienced players under the evaluated conditions; it is not an estimate of population-level human performance.
\section{Experiments and Analysis}
\label{sec:experiments}
\label{sec:analysis}
\subsection{Experimental Setup}
We evaluate GPT-6 Astra, GPT-6.1 Sol, and DeepSeek-v4.1-Flash at low, medium, and high reasoning effort. Each of the nine configurations is evaluated on all 36 tasks, yielding 324 task runs and 17,221 completed model responses. Appendix~\ref{app:agent-config} specifies the prompts, history retention, and generation parameters.

\label{sec:protocol}
\paragraph{Execution budgets.}
Each episode uses its task's time budget $B_i$ and a maximum of 100 model responses; one response may contain multiple actions. The runner checks the budget before each request and each action. An in-flight request or action is not forcibly interrupted. Each evaluator uses its specified evidence: a deadline snapshot, timestamp-filtered events, or the final state after the interaction loop. Appendix~\ref{app:action-termination} details stopping conditions.

\paragraph{Metrics.}
For $N=36$ tasks, let $Y_i$ be the binary task outcome and $z_{ij}=1$ when subgoal $j$ is verified as achieved, and zero otherwise. We report
\begin{equation}
\mathrm{Task\ success\ rate}=\frac{1}{N}\sum_{i=1}^{N}Y_i,\qquad
\mathrm{Subgoal\ success\ rate}=\frac{1}{N}\sum_{i=1}^{N}\frac{1}{m_i}\sum_{j=1}^{m_i}z_{ij}.
\label{eq:goals}
\end{equation}
Both rates are displayed as percentages. Each task receives equal weight; completing one subgoal does not count as completing a task. Initialization failures receive no success credit and remain in the denominator. Appendix~\ref{app:subgoal-diagnostics} reports subgoal breakdowns and analyses that pool individual checks.

We also report mean completed-request latency and mean responses per executed episode. Request latency includes service, network, retry, and parsing overhead, so it measures response delay rather than reasoning time alone. Together, these measures relate task completion and partial progress to the time and decision opportunities available to the agent.

\subsection{Overall Results}
As shown in Table~\ref{tab:main} and Figure~\ref{fig:capabilities}, GPT-6 Astra at low and medium effort shares the highest task success rate, with 9/36 passes each (25.0\%). Across the cohort, 40/324 runs pass (12.3\%) and 284 fail, covering 12 distinct passing tasks. Economy and development each account for 19 passes. The remaining two passes are on Malphite's combat-category task, which requires reaching level~5 with at most one death rather than securing a kill. Sustain, cooperation, and map-objective tasks have no successful run.

\begin{table}[ht]
\centering\small
\caption{\textbf{Task success rate and subgoal success rate under a common evaluator.} Task success rate is confirmed full-task passes divided by 36; subgoal success rate is the mean fraction of successful subgoals per task. All configurations use the same task evaluators. Turns and latency describe recorded execution. Bold marks highest scores; comparisons are descriptive.}
\label{tab:main}
\setlength{\tabcolsep}{3pt}
\renewcommand{\arraystretch}{1.06}
\begin{tabular*}{\linewidth}{@{\extracolsep{\fill}}llrrrr@{}}
\toprule
\textbf{Model} & \textbf{Effort} & \textbf{\shortstack{Task success\\rate (\%)}} & \textbf{\shortstack{Subgoal success\\rate (\%)}} & \textbf{Mean Turns} & \textbf{Mean Latency (s)}\\
\midrule
\multirow{3}{*}{GPT-6 Astra} & Low & \textbf{25.0} & 37.7 & 43.9 & 7.47 \\
 & Medium & \textbf{25.0} & \textbf{40.3} & 37.8 & 8.39 \\
 & High & 16.7 & 33.7 & 27.1 & 16.67 \\
\midrule
\multirow{3}{*}{GPT-6.1 Sol} & Low & 13.9 & 32.1 & 42.8 & 8.90 \\
 & Medium & 11.1 & 31.3 & 33.2 & 11.44 \\
 & High & 11.1 & 24.0 & 37.4 & 10.26 \\
\midrule
\multirow{3}{*}{DeepSeek-v4.1-Flash} & Low & 5.6 & 22.2 & 89.8 & 2.81 \\
 & Medium & 2.8 & 20.0 & 91.9 & 3.05 \\
 & High & 0.0 & 18.1 & 77.0 & 4.67 \\
\bottomrule
\end{tabular*}
\end{table}

Using the same task-averaged metric as Table~\ref{tab:main}, subgoal success rate is 28.8\% across all nine configurations. GPT-6 Astra (medium) achieves the highest subgoal success rate (40.3\%), above low effort's 37.7\%, while both complete nine tasks. Across 284 failed runs, 108 (38.0\%) achieve some required subgoals without completing the task. Thus additional subgoal attainment need not increase the number of completed tasks. Appendix~\ref{app:subgoal-diagnostics} gives additional subgoal analysis; Figure~\ref{fig:matrix} reports every task and configuration.

\subsection{Subgoal Analysis}
\begin{figure}[!htbp]
\centering
\includegraphics[width=\linewidth]{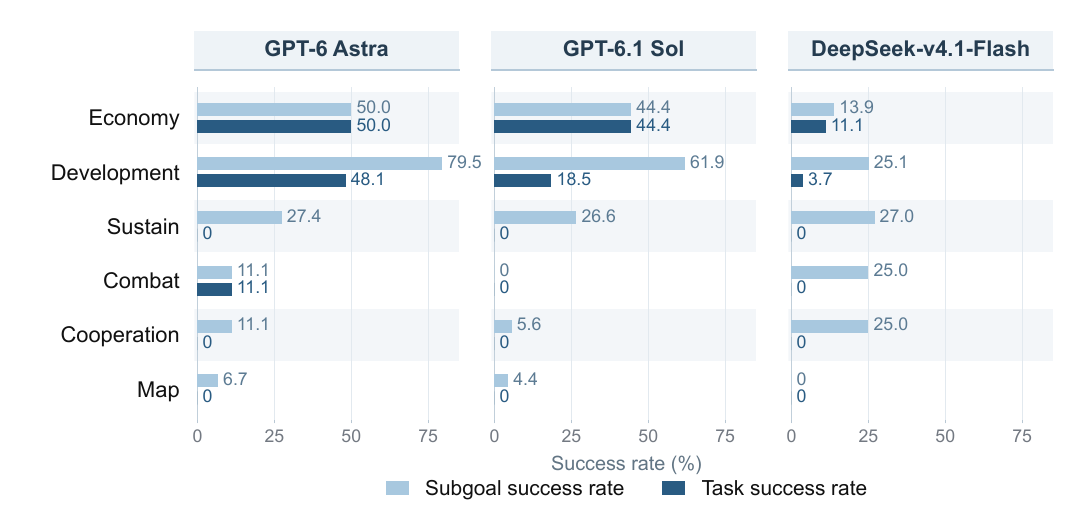}
\caption{\textbf{Partial progress rarely becomes full task success.} Paired bars compare subgoal success rate with task success rate (\%), pooling three effort settings per model. All model panels share the same scale and category rows; lighter and darker bars indicate subgoal success rate and task success rate, respectively. Three families have zero full success for every model.}
\label{fig:capabilities}
\end{figure}
Figure~\ref{fig:capabilities} shows where subgoal success fails to translate into task success. Development tasks, which require ability allocation or attribute growth, average a 55.5\% subgoal success rate but only a 23.5\% task success rate; sustain tasks, which require maintaining health, mana, or survival, average a 27.0\% subgoal success rate with no successful episode. A task still fails if one required subgoal is unmet or its timing requirement is violated, even when other subgoals succeed.

Breaking down success by the requirement being checked reveals which parts of a task agents can complete. Checks for minimum learned ability levels are achieved in 25/36 instances and level thresholds in 52/108, compared with 4/9 checks for possession of the required item, 10/144 checks for a minimum creep score (CS, the count of minion and monster last hits), 0/45 personal-kill checks, and 0/27 allied-assist checks. The contrast is strongest between learning abilities or reaching a level and completing combat objectives: agents satisfy some character-development requirements, yet no evaluated run satisfies the personal-kill or allied-assist checks.

Some subgoals require making progress, such as buying an item; others require preserving a condition, such as staying alive or leaving an ability unlearned. Satisfying the latter does not necessarily demonstrate successful combat or navigation. This distinction helps explain why partial subgoal success can coexist with task failure. Appendix~\ref{app:subgoal-diagnostics}, Table~\ref{tab:subgoal-semantics}, defines these groups and reports their counts.

\subsection{Latency and Effective Interaction}
\begin{figure}[!htbp]
\centering
\includegraphics[width=\linewidth]{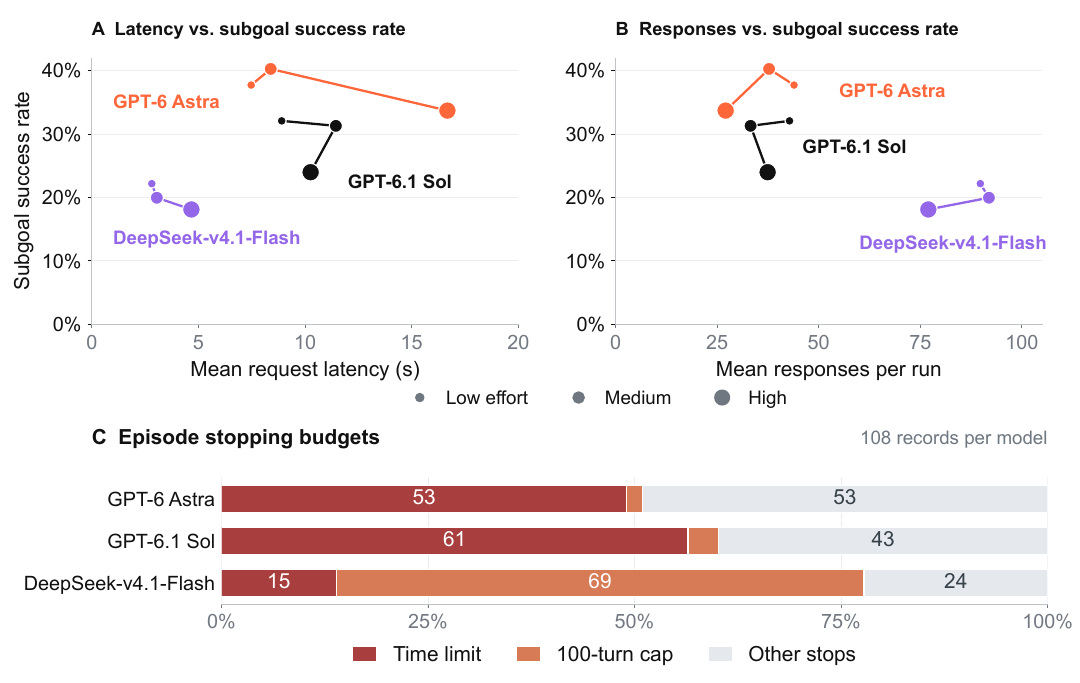}
\caption{(A) Subgoal success rate versus mean completed-request latency. (B) Subgoal success rate versus mean responses per executed run. (C) Distribution of stopping conditions by model, pooled across reasoning-effort settings.}
\label{fig:temporal}
\end{figure}
\textbf{Lower latency gives agents more chances to adapt.} While a model generates its response, enemies move, combat unfolds, and interface states can change. Shorter requests bring the next observation forward, allowing the agent to detect these changes and revise its actions sooner. For GPT-6 Astra, low effort reduces mean request latency from 16.67 seconds at high effort to 7.47 seconds, while increasing mean responses per episode from 27.1 to 43.9 (Figure~\ref{fig:temporal}). Task success also rises from 16.7\% to 25.0\%. This within-model pattern is consistent with faster feedback helping the agent adapt to an evolving scene and complete more tasks within the same time budget.

\textbf{Accurate visual grounding makes fast responses useful.} Across models, DeepSeek-v4.1-Flash (low) averages 2.81 seconds per request and 89.8 responses per executed episode, compared with 7.47 seconds and 43.9 responses for GPT-6 Astra (low), yet achieves lower task success (5.6\% versus 25.0\%) and subgoal success (22.2\% versus 37.7\%). The side-by-side Lee Sin demonstration\footnote{Side-by-side gameplay: \url{https://answers111.github.io/LOL-Play-homepage/\#task002-video}.} illustrates how this speed advantage can be lost. DeepSeek-v4.1-Flash responds more frequently but repeatedly clicks at incorrect coordinates, spending many steps in the shop interface without completing the intended interaction. It reaches the 100-step cap at level~1 with zero CS; GPT-6 Astra reaches level~3 and 8 CS before its time limit. Recognizing interface elements and mapping intended targets to the correct pixels is therefore necessary to turn frequent responses into useful control. The observed off-target clicks identify a visual-grounding failure in this example.

\textbf{Ineffective actions consume the execution budget.} In the benchmark, DeepSeek-v4.1-Flash at low and medium effort reaches the 100-response ceiling in 26 and 29 tasks, respectively, but the time limit in only one and two; GPT-6 Astra at high effort exhausts its time budget in 23 tasks (Figure~\ref{fig:temporal}C). Of DeepSeek's 55 capped low/medium runs, 32 achieve no subgoal, and only 5/93 positive-attainment checks pass across all 55 runs. The other 23 achieved checks concern avoidance or state preservation (Appendix~\ref{app:response-cap}). These aggregate results establish limited progress despite frequent interaction; the side-by-side gameplay video provides one concrete example of how inaccurate clicks can waste those opportunities. Effective real-time control requires both timely responses and actions that correctly affect the intended targets.

\subsection{Reasoning Effort}
\begin{figure}[!htbp]
\centering
\includegraphics[width=\linewidth]{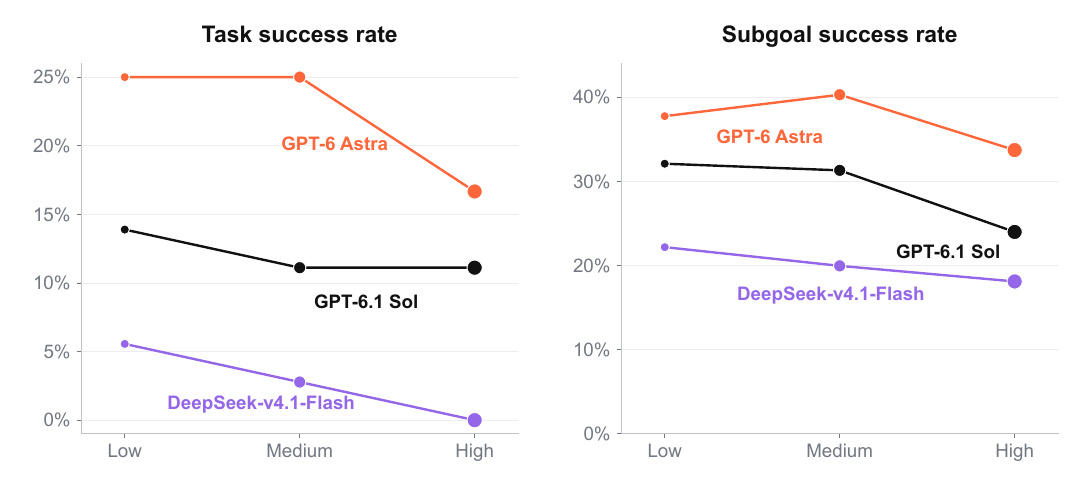}
\caption{\textbf{High effort does not improve success over low effort.} Task success rate (left) and subgoal success rate (right) over 36 tasks. Lines connect low, medium, and high effort settings. Astra ties in task success at low and medium effort, with higher subgoal success at medium effort.}
\label{fig:effort}
\end{figure}
Low effort achieves or ties the highest task success rate for each evaluated model (Figure~\ref{fig:effort}). Astra completes 9, 9, and 6 tasks at low, medium, and high effort (25.0\%, 25.0\%, and 16.7\%); Sol completes 5, 4, and 4 (13.9\%, 11.1\%, and 11.1\%); DeepSeek-v4.1-Flash completes 2, 1, and 0 (5.6\%, 2.8\%, and 0\%). Increasing effort therefore yields no task-success gain in these evaluated configurations.

Subgoal success distinguishes gains in partial progress from completed tasks. GPT-6 Astra (medium) reaches 40.3\%, above low's 37.7\%, while high falls to 33.7\%. Sol declines from 32.1\% to 31.3\% to 24.0\%, and DeepSeek-v4.1-Flash from 22.2\% to 20.0\% to 18.1\%. Thus medium effort benefits Astra's subgoal attainment without increasing its count of completed tasks; high effort improves neither metric over low effort for any model.

These results motivate testing how additional reasoning interacts with real-time execution. Astra's low-to-high latency more than doubles while task success falls by one-third, but Sol's latency is non-monotonic (8.90, 11.44, and 10.26 seconds). Effort is therefore not a uniform latency intervention, and the results do not isolate a causal effect of delay. The observed tradeoff supports evaluating reasoning effort jointly by task completion, subgoal attainment, and response latency rather than assuming that more reasoning improves real-time control.

\subsection{Case Study}
\begin{figure}[!htbp]
\centering
\includegraphics[width=\linewidth]{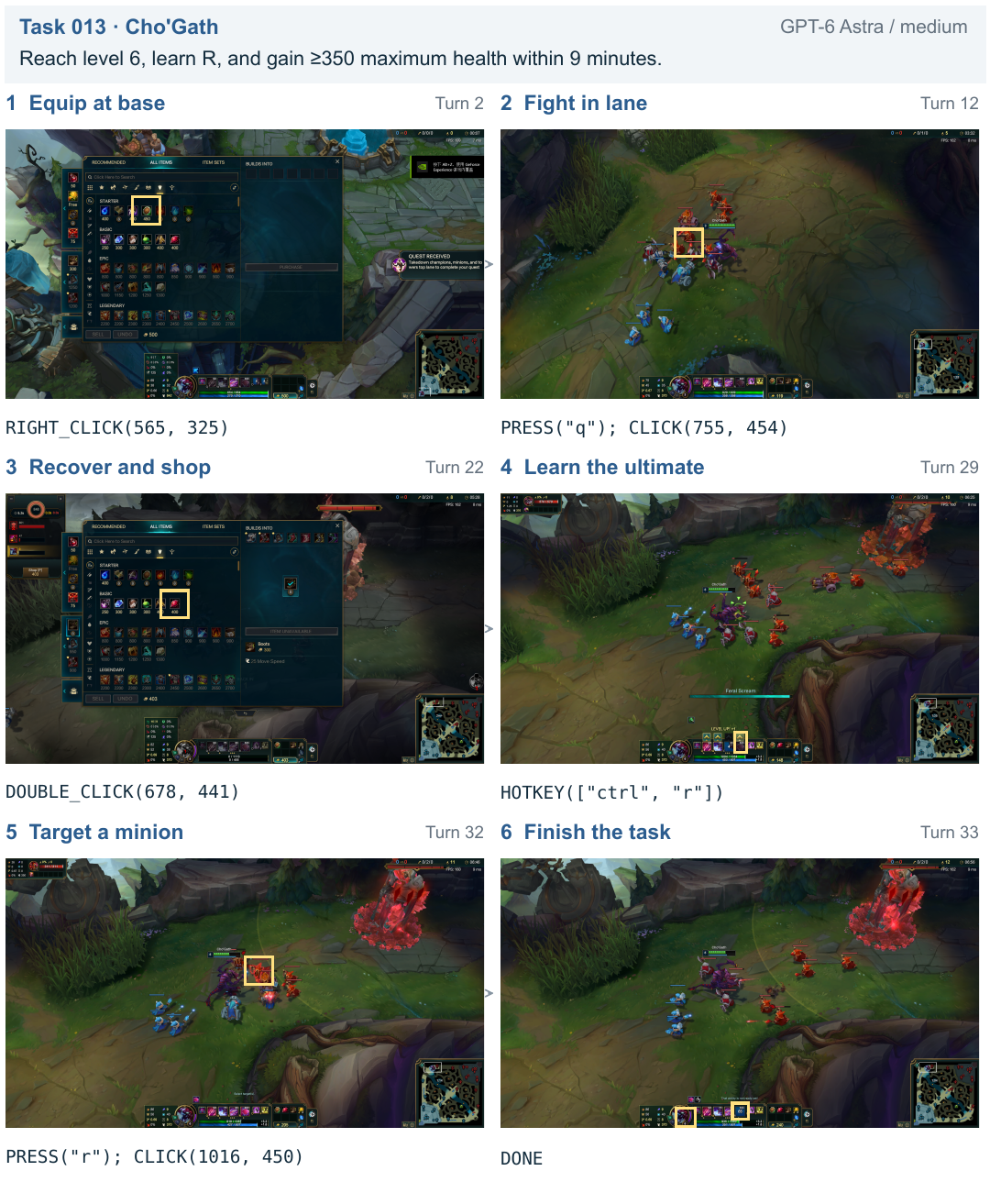}
\caption{\textbf{From task instruction to verified completion.} Six selected turns from a successful Cho'Gath episode, read left to right and top to bottom. Each image shows the observation used for the labeled decision; the command below is an excerpt from that turn's executed-action log, with JSON keys omitted. Yellow outlines highlight click targets, ultimate allocation, and final level/skill indicators. Intermediate turns and other actions are omitted.}
\label{fig:success-walkthrough}
\end{figure}

\paragraph{Completing a task.}
\label{sec:success-case}
Figure~\ref{fig:success-walkthrough} follows a successful GPT-6 Astra (medium) run on task~013. Cho'Gath must reach level~6, learn R, and increase maximum health by at least 350 within nine minutes. Six selected decision inputs show equipment purchase, lane combat, recovery and shopping, ultimate allocation, a targeted R command, and the final \code{DONE} signal. The full episode contains 33 responses and two deaths: success here means satisfying the task contract, not uninterrupted or optimal play.

The final evidence confirms all three checks: level~6, R rank~1, and a maximum-health increase of 795.3, after 420.8 seconds of execution against a 540-second budget. This trajectory shows how generic desktop inputs can span shopping, movement, combat, and skill allocation within one successful task. It also clarifies the evaluator's scope: the checks establish the final state, not the success of every attempted cast or the share of health growth attributable to Feast rather than levels and items. The next case contrasts this completed task with local progress that does not become task success.

\begin{figure}[!htbp]
\centering
\includegraphics[width=\linewidth]{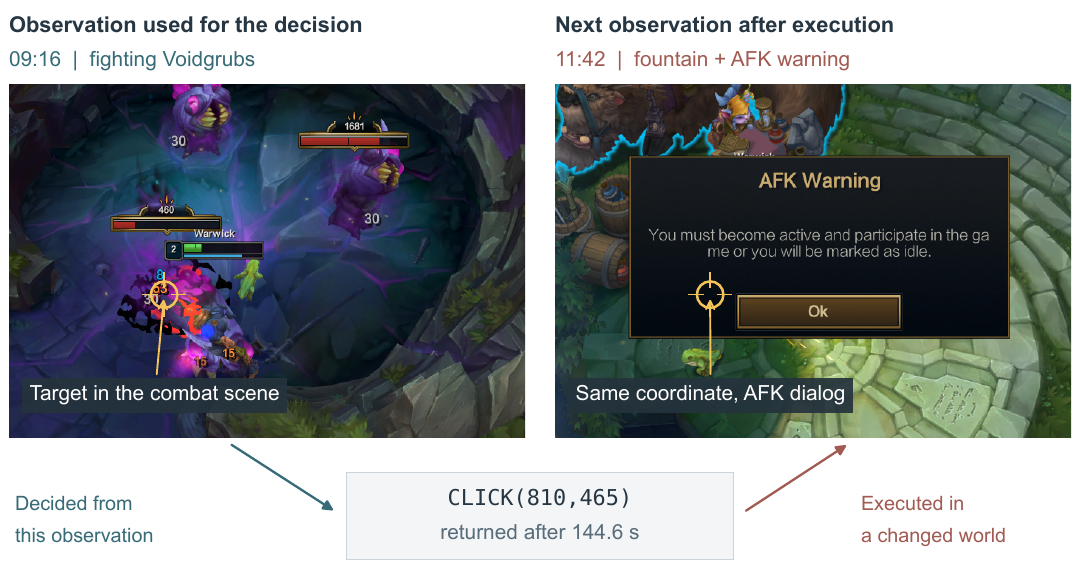}
\caption{\textbf{Decisions use an old observation; actions execute in a changed world.} One logged click in task 035 (GPT-6 Astra (medium)) links its input observation to the next post-execution frame, after a 144.641-second request. Both crosshairs mark (810,465): a combat scene becomes an AFK dialog. The right frame follows the full action batch; exact per-action states and input interception are not recorded.}
\label{fig:actiontrace}
\end{figure}

\paragraph{Acting on a stale observation.}
\label{sec:stale-case}
Figure~\ref{fig:actiontrace} traces task 035 (GPT-6 Astra (medium)), in which Warwick must personally kill a Voidgrub and then destroy the enemy top outer turret.

\paragraph{Observation delay.}
At 09:16, the input shows Warwick fighting Voidgrubs. During the 144.641-second request, game events show a personal Voidgrub kill at 09:19.65 and a death at 09:31.74. The next frame, captured at 11:42 after the action batch, shows the champion at the fountain behind an inactivity warning. The encounter has ended before the response arrives.

\paragraph{Action mismatch.}
The agent executes \code{CLICK(810,465)}. This coordinate refers to the combat scene in the input but lies inside the AFK dialog in the next frame. A fresh observation would therefore need to account for both a changed game state and an obstructing interface. The two frames do not establish when the dialog appeared, which inputs it intercepted, or the effects of individual actions.

\paragraph{Task outcome.}
Only one of three checks passes: the personal Voidgrub kill. The turret remains intact and the follow-up sequence is incomplete. This case links an outdated observation to actions executed in a changed context, while showing why a local achievement need not yield task success. It motivates refreshing observations and replanning after intervening events; it does not establish that latency alone caused death or failure.

\paragraph{What the cases reveal.}
Together, these trajectories distinguish four aspects of control. \textbf{Perception:} Warwick's input depicts an encounter that ends during inference; this demonstrates stale information, rather than a demonstrated misreading of the input image. \textbf{Click grounding:} the same screen coordinate refers to different content in the two frames, exposing a change in the action's context rather than proving that its original location was incorrectly predicted. \textbf{Planning:} Warwick completes the first objective but not the required follow-up, while Malphite preserves survival without reaching the target level. These are identifiable gaps in goal progression, although the records do not uniquely attribute them to planning. \textbf{Execution:} logged commands establish which actions were issued; the Cho'Gath outcome verifies complete task success, whereas the Warwick frames do not establish which inputs the dialog intercepted. The cases therefore explain distinct observable breakdowns without treating them as mutually exclusive error labels. They complement the aggregate subgoal results; they neither estimate error frequencies nor replace a fixed-model delay intervention.

\section{Related Work}
\label{sec:related}
\paragraph{Computer-use agents and evaluation.}
Mind2Web~\cite{mind2web} provides demonstrations across real websites, while WebVoyager~\cite{webvoyager} studies end-to-end multimodal web interaction. WebArena~\cite{webarena} and VisualWebArena~\cite{visualwebarena} evaluate functional website tasks; WorkArena~\cite{workarena} and WorkArena++~\cite{workarenapp} examine enterprise workflows and their composition. AndroidWorld~\cite{androidworld}, OSWorld~\cite{osworld}, and Windows Agent Arena~\cite{windowsarena} extend executable evaluation to mobile and desktop applications. OSWorld~2.0~\cite{osworld2} adds long workflows, functional checkpoints, and dynamic interfaces. SeeClick~\cite{seeclick} studies screenshot-based GUI grounding, and ScreenSpot-Pro~\cite{screenspotpro} tests target localization in high-resolution interfaces. \benchmark{} connects visual grounding and multi-step interaction to goals whose execution unfolds in a continuously changing MOBA.

\paragraph{Real-time interaction and latency.}
OSWorld-Human~\cite{osworldhuman} evaluates elapsed-time efficiency. VideoGameBench~\cite{videogamebench} studies whole-game progress through predefined checkpoints and contrasts real-time play with paused Lite episodes. It already identifies latency as a bottleneck and examines perception, planning, and basic control. \benchmark{} contributes a different evaluation unit: champion-specific objectives with jointly required state, participant, and event-order checks. These reveal whether partial achievements satisfy the requested objective, rather than establishing latency as a new phenomenon. GameWorld~\cite{gameworld} supports paused and real-time protocols; Realtime-Bench~\cite{realtimebench} compares clock protocols; ROAR~\cite{roar} decouples updates from inference. Our study complements these temporal analyses with task and subgoal outcomes in a native MOBA.

\paragraph{General-purpose game agents.}
GameBench~\cite{gamebench} evaluates strategic reasoning across games; BALROG~\cite{balrog} and Orak~\cite{orak} assess broader game-playing capabilities, while GameVerse~\cite{gameverse} studies video-based reflection and retry. Cradle~\cite{cradle} explores screenshot-driven control in commercial games, and SIMA~\cite{sima} and NitroGen~\cite{nitrogen} develop generalist visual-action policies. Minecraft supports complementary approaches: MineRL~\cite{minerl} supplies human demonstrations, VPT~\cite{vpt} learns action policies from video, MineDojo~\cite{minedojo} supports open-ended tasks, and Voyager~\cite{voyager} builds reusable skills. Hierarchical Language Agents~\cite{hla} separate reasoning and reactive execution in Overcooked. \benchmark{} evaluates existing agents through champion-specific goals, exposing which requirements are achieved and which combinations remain incomplete.

\paragraph{Game environments and specialized agents.}
The Arcade Learning Environment~\cite{ale} supports evaluation across Atari games. SC2LE~\cite{sc2le} provides StarCraft~II maps and skill-focused mini-games; Crafter~\cite{crafter} measures progress through achievements in a visual survival environment. Thus, component-level game evaluation predates \benchmark{}; our contribution is its task-level conjunction with uninterrupted native MOBA interaction. OpenAI Five~\cite{openaifive}, AlphaStar~\cite{alphastar}, and full-game MOBA reinforcement learning~\cite{fullmoba} demonstrate sophisticated control with specialized training. \benchmark{} evaluates general-purpose agents through screenshots and desktop actions against explicit user objectives. Its task scores do not imply competitive strength against trained esports agents.

\section{Conclusion and Discussion}
\label{sec:limitations}
\paragraph{Conclusion.}
\benchmark{} evaluates real-time visual agents through 36 champion-specific tasks in a native MOBA. The highest task success rate is 25.0\%, and 108 failed runs achieve partial progress. Astra's medium effort improves subgoal success over low effort without completing more tasks, while high effort combines longer requests with fewer completed tasks. Together with the stale-observation case study, these findings motivate agents that coordinate reasoning with timely observation and action while satisfying multiple requirements.

\paragraph{Limitations.}
The 36 tasks of LOLPlay cover one objective per champion rather than the full diversity of MOBA play. Some requirements can be satisfied through passive progress, making task success essential alongside subgoal success. These boundaries limit the interpretation of the scores without changing the observed outcomes.
The intended setting is custom matches with bots. Game assets belong to Riot Games. Desktop input and unrestricted Python require a trusted environment; uncontrolled deployment could enable unwanted gameplay automation.

\bibliographystyle{plainnat}
\bibliography{bibliography}
\clearpage
\appendix
\section{Task and Evaluation Details}
\label{app:system}
This appendix supplements the task definitions, agent interface, environment setup, and evaluation in Section~\ref{sec:benchmark}.

\subsection{Complete Task List}
\label{app:task-contract}
\label{app:catalog}
Table~\ref{tab:catalog} lists all 36 task objectives and time budgets. Each task belongs to one primary family for reporting, although its requirements can involve several skills. The 82 subgoal checks count each required item, ability rank, enemy victim, and turret separately; an ordering requirement adds a check. Match validity and the overall time limit are evaluated separately from these subgoals.
\begingroup
\small\setlength{\tabcolsep}{3pt}
\begin{longtable}{@{}p{.29in}p{.78in}p{.26in}p{3.90in}@{}}
\caption{\textbf{Complete task objectives.} Budgets are nominal minutes. Numeric item IDs match the evaluator, which checks inventory membership rather than build ancestry. Context validity and global timing also apply. All reported configurations use these targets.}\label{tab:catalog}\\
\toprule ID & Champion & Min. & Numbered outcome checks \\ \midrule
\endfirsthead
\toprule ID & Champion & Min. & Numbered outcome checks (continued) \\ \midrule
\endhead
\bottomrule\endfoot
001 & Caitlyn & 10 & CS $\geq 30$ \\
002 & Lee Sin & 10 & personally kill first elemental dragon \\
003 & Ziggs & 10 & destroy enemy mid outer turret \\
004 & Sion & 10 & max HP $\geq 2000$ \\
005 & Varus & 7 & CS $\geq 10$ \\
006 & Mordekaiser & 8 & CS $\geq 10$ \\
007 & Sona & 8 & level $\geq 4$; gold $\geq 300$ \\
008 & Rammus & 8 & level $\geq 3$ \\
009 & Ahri & 7 & CS $\geq 10$; level $\geq 4$; rank Q $= 2$; rank W $= 1$; rank E $= 1$; rank R $= 0$ \\
010 & Kayle & 9 & CS $\geq 10$; level $\geq 6$; rank R $\geq 1$ \\
011 & Garen & 7 & CS $\geq 10$; level $\geq 4$; rank Q $\geq 1$; rank E $\geq 1$; rank W $= 0$ \\
012 & Veigar & 8 & CS $\geq 10$; AP gain $\geq 40$ \\
013 & Cho'Gath & 9 & level $\geq 6$; rank R $\geq 1$; max HP gain $\geq 350$ \\
014 & Ashe & 8 & CS $\geq 20$; attackSpeed $\geq 0.9$ \\
015 & Leona & 8 & level $\geq 4$ \\
016 & Brand & 8 & CS $\geq 20$; mana fraction $\geq 0.5$ \\
017 & Dr. Mundo & 8 & CS $\geq 20$; health fraction $\geq 0.8$; require alive \\
018 & Soraka & 8 & CS $\leq 19$; level $\geq 5$ \\
019 & Lux & 8 & CS $\geq 20$; rank W $= 0$ \\
020 & Kennen & 8 & CS $\geq 20$; level $\geq 4$; deaths $\leq 0$ \\
021 & Sivir & 8 & CS $\geq 10$; deaths $\leq 0$ \\
022 & Darius & 9 & deaths $\leq 1$; personal kills $\geq 1$ \\
023 & Tristana & 9 & CS $\geq 10$; personal kills $\geq 1$ \\
024 & Malphite & 9 & level $\geq 5$; deaths $\leq 1$ \\
025 & Lulu & 10 & deaths $\leq 1$; min assists $\geq 1$ \\
026 & Nautilus & 10 & level $\geq 5$; min assists $\geq 2$ \\
027 & Yorick & 20 & turret top / outer; turret top / inner; in required order \\
028 & Jax & 20 & turret top / outer; turret middle / outer \\
029 & Annie & 10 & personal kills $\geq 2$; deaths $\leq 1$ \\
030 & Talon & 5 & personal kills $\geq 1$; deaths $\leq 0$ \\
031 & Katarina & 10 & personally kill Miss Fortune; personally kill Soraka; deaths $\leq 2$ \\
032 & XinZhao & 12 & participations $\geq 3$; personal kills $\geq 1$; min ally assists $\geq 1$; deaths $\leq 2$ \\
033 & Draven & 10 & CS $\geq 40$; min current gold $\geq 1500$; alive \\
034 & Ryze & 10 & CS $\geq 40$; level $\geq 7$; mana fraction $\geq 0.6$; alive \\
035 & Warwick & 12 & personal Voidgrubs $\geq 1$; turret top / outer; in required order \\
036 & Syndra & 7 & item 3802 $\geq 1$ \\
\end{longtable}
\endgroup

\subsection{Action Parameters and Termination}
\label{app:observation-contract}
\label{app:action-termination}
Screenshots cover the primary display, with a foreground-window check before observation and input. Image resolution is not fixed by the interface; the case-study images are $1920\times1080$ pixels. The agent completes one observation--response--action cycle before starting the next. Actions execute after a complete response is received and parsed, while gameplay continues throughout.

Table~\ref{tab:action-space-full} gives the action parameters supplementing Table~\ref{tab:action-space}. Coordinates use absolute screenshot pixels, and actions in a response execute sequentially. Table~\ref{tab:terminal-states-full} lists stopping conditions and their observed counts separately from task success.
\begin{table}[htbp]
\centering
\small
\setlength{\tabcolsep}{4pt}
\renewcommand{\arraystretch}{1.12}
\caption{\textbf{Complete structured action space.} The \code{computer\_13} adapter supports 13 mouse--keyboard action types and three control markers. Coordinates are absolute screenshot pixels. A response may contain multiple actions, executed sequentially; the game remains live throughout.}
\label{tab:action-space-full}
\begin{tabularx}{\linewidth}{@{}>{\raggedright\arraybackslash}p{.14\linewidth}>{\raggedright\arraybackslash}p{.40\linewidth}>{\raggedright\arraybackslash}X@{}}
\toprule
Group & Actions & Arguments and behavior \\
\midrule
Pointer & \code{MOVE\_TO}, \code{DRAG\_TO} & Target $(x,y)$; optional duration. Move the pointer or drag with the left button. \\
Click & \code{CLICK}, \code{RIGHT\_CLICK}, \code{DOUBLE\_CLICK} & Optional $(x,y)$; otherwise use the current pointer. \code{CLICK} also accepts a button and 1--3 clicks. \\
Mouse hold & \code{MOUSE\_DOWN}, \code{MOUSE\_UP} & Press or release a button (left by default). \\
Scroll & \code{SCROLL} & Integer horizontal and/or vertical displacement, \code{dx}/\code{dy}. \\
Keyboard & \code{PRESS}, \code{KEY\_DOWN}, \code{KEY\_UP} & Key name; press, hold, or release. \code{PRESS} accepts an optional duration. \\
Shortcut & \code{HOTKEY} & A list of keys forming a key combination. \\
Text & \code{TYPING} & Text string; optional typing interval and key-press duration. \\
\midrule
Wait & \code{WAIT} & Wait for the agent-configured interval; continue the episode afterward. \\
Stop & \code{DONE}, \code{FAIL} & Declare completion or inability to proceed; stop the agent loop and obtain an independent evaluation. \\
\bottomrule
\end{tabularx}
\end{table}

\begin{table}[htbp]
\centering
\small
\setlength{\tabcolsep}{4pt}
\renewcommand{\arraystretch}{1.1}
\caption{\textbf{Complete stopping reasons and final outcomes.} Counts describe the 324-run cohort. Each block partitions the same records independently. The two zero-count runner branches are supported but were not observed. A \code{DONE} signal, a time limit, or a turn limit does not determine evaluator success.}
\label{tab:terminal-states-full}
\begin{tabularx}{\linewidth}{@{}>{\raggedright\arraybackslash}p{.37\linewidth}>{\raggedright\arraybackslash}Xr@{}}
\toprule
Status & Meaning & Count \\
\midrule
\multicolumn{3}{@{}l}{\textit{Runner stopping reason}} \\
\code{done} & Agent emits \code{DONE}. & 37 \\
\code{infeasible} & Agent emits \code{FAIL}. & 60 \\
\code{execution\_budget\_exhausted} & The wall-time/task stop condition is reached. & 129 \\
\code{max\_steps} & The model-response cap is reached (100 here). & 75 \\
\code{execution\_error} & An exception interrupts prediction or execution. & 22 \\
\code{setup\_error} & Preparation fails before the agent loop. & 1 \\
\code{no\_actions} & Prediction returns an empty action list. & 0 \\
\code{environment\_terminated} & An environment step reports termination. & 0 \\
\midrule
\multicolumn{3}{@{}l}{\textit{Final outcome under the common evaluation protocol}} \\
Task success & The evaluator confirms task completion. & 40 \\
Failure & Task requirements are not met, or setup fails. & 284 \\
\bottomrule
\end{tabularx}
\end{table}

\subsection{Setup and Evaluation APIs}
\label{app:lcu-interfaces}
The following API operations expand the setup and evaluation interfaces in Table~\ref{tab:api-interfaces}. LCU configures the game client; the separate Live Client Data service provides in-game state for evaluation.
\begin{itemize}
\item \textbf{Lobby and bots.} Create the match with \code{POST /lol-lobby/v2/lobby}; query available bots, then add the configured champion, difficulty, and team through \code{POST /lol-lobby/v1/lobby/custom/bots}. Read back the roster before starting selection through \code{/lol-lobby/v1/lobby/custom/start-champ-select}.
\item \textbf{Champion and loadout.} Read the champion-selection session and complete the controlled player's pick action. Update summoner spells and skin through the selection service's \code{session/my-selection} endpoint. Configure an editable rune page through \code{/lol-perks/v1/pages} and activate it through \code{/lol-perks/v1/currentpage}.
\item \textbf{Input settings and readiness.} Read and patch \code{/lol-game-settings/v1/input-settings}, save through \code{/lol-game-settings/v1/save}, and verify normal casting. Poll \code{/lol-gameflow/v1/gameflow-phase} until the game begins.
\item \textbf{Evidence.} Query \code{GET /liveclientdata/allgamedata} from the separate in-game service. Extract state checks and attributed, timestamped events; apply task-start, deadline, and ordering conditions where required. This service is used for validation and evaluation, not to execute agent actions.
\end{itemize}

\subsection{Success Checks and Timing}
\label{app:timing-contract}
Table~\ref{tab:lcd-checkers} maps Live Client Data fields to the task checks illustrated in Table~\ref{tab:task-examples}.
\begin{table}[htbp]
\centering
\small
\setlength{\tabcolsep}{4pt}
\renewcommand{\arraystretch}{1.12}
\caption{\textbf{Live Client Data fields used by task checkers.} All fields are read through \code{GET /liveclientdata/allgamedata}; setup retains required baselines. A check receives success credit only when the required evidence is available.}
\label{tab:lcd-checkers}
\begin{tabularx}{\linewidth}{@{}p{.12\linewidth}p{.38\linewidth}X@{}}
\toprule
Evidence & Fields / payload group & Checker use \\
\midrule
Context & \code{gameData}: \code{gameTime}, \code{gameMode}, \code{mapNumber} & Validate match context and elapsed game time against task-start time and budget. \\
Player & \code{allPlayers}: champion, identity, \code{team}, \code{level}, \code{isDead} & Match the controlled player and roster; check level and survival; attribute events to teams. \\
Scores & \code{scores}: \code{creepScore}, \code{deaths}, \code{kills}, \code{assists} & Farming and death constraints; retain score evidence. Attributed combat goals additionally use events. \\
Inventory & \code{items}: \code{itemID}, \code{count}; \code{activePlayer.currentGold} & Check required item identities/counts and available gold. \\
Abilities & \code{activePlayer.abilities}: Q/W/E/R \code{abilityLevel} & Check learned or unlearned abilities and required ranks. \\
Attributes & \code{activePlayer.championStats}: health, resource, and target attributes & Check health/mana ratios, maximum health, and attribute gain relative to the setup baseline. \\
Events & \code{events.Events}: \code{EventID}, \code{EventName}, \code{EventTime}; killer, victim, assisters, objective fields & Validate event identity/time; attribute kills and assists; check dragon type, turret lane/tier, Voidgrub ownership, and required order. \\

\bottomrule
\end{tabularx}
\end{table}

\paragraph{Creep score.}
The evaluation accounts for observed ten-point updates in client-reported creep score, modeled as $\widehat{c}=10\lfloor c/10\rfloor$. Minimum-CS thresholds use multiples of ten, and the low-CS requirement uses $c<20$ rather than $c\leq10$. The evaluator uses the API value without inferring a precise count inside a bin. This behavior was observed in the evaluated client and is not assumed to hold across future updates. CS includes both minion and neutral-monster last hits.

\paragraph{Interpreting success checks.}
Turret checks establish destruction of the specified target and accept allied-minion last hits. Attribute gains may arise from levels, items, or champion abilities; current gold measures the unspent balance. Ability ranks verify allocation rather than casting proficiency. Assist checks establish participation in an allied-bot kill. These are outcome checks and do not establish the quality of each intervening action.

\subsection{Agent Configuration}
\label{app:agent-config}
All nine evaluated configurations use the same screenshot-to-action interface and prompt template, varying the model and reasoning-effort setting. 

\paragraph{Request construction.}
Each request supplies the system prompt below, followed by the literal header \code{Task instruction:} and the task's instruction text. The task instruction is repeated on every request. The user message contains the screenshot and the text ``Inspect this screenshot and return the next computer\_13 JSON action or special marker.'' Previous screenshot messages and assistant response texts are interleaved in chronological order before this message. History is reset at the start of each task and truncated to the latest three valid turns; it is not summarized. Separate hidden reasoning items and game-state evaluation data are not added to this history.

\paragraph{System prompt.}
The following reproduces the prompt text, including its formatting examples. The single placeholder denotes the action schema supplied in the actual prompt: its actions and arguments are summarized in Table~\ref{tab:action-space-full}, and the complete machine-readable schema is included in the released prompt template. It is elided here to avoid duplicating that table; the model receives the schema, not the placeholder.
\begin{tcolorbox}[breakable,colback=black!3,colframe=black!65,
  colbacktitle=black!65,coltitle=white,fonttitle=\bfseries\small,
  title={System prompt: screenshot-to-action agent},
  title after break={System prompt (continued)},
  boxrule=0.5pt,arc=1mm,left=7pt,right=7pt,top=6pt,bottom=6pt,
  before skip=10pt,after skip=10pt]
\begin{Verbatim}[fontsize=\small,breaklines=true,breakanywhere=true]
You will act as an agent which follows my instructions and performs League of Legends tasks on Windows.
For each step, you will get an image observation: a screenshot of the computer screen.
Predict the next computer action from this screenshot.

Here is the action space. Follow its action_type names and parameter definitions:
[Action schema omitted here; see the complete action-space table.]

Parameter types above use JSON schema notation: number, integer, string and array.
A null range means no additional range restriction; a null upper bound means unbounded.
X_MAX and Y_MAX correspond to the current screenshot dimensions, not a fixed 1920x1080 screen.
Coordinates are absolute screenshot pixels: the top-left corner is (0, 0).
Whenever coordinates are optional, provide both x and y or neither.
SCROLL must include at least one of dx and dy. Use the supported key names from
pydirectinput-rgx, such as q, w, e, r, ctrl, shift, alt, enter, esc, space and arrow keys.
Key aliases such as ArrowUp are accepted and normalized.

First choose an action_type from ACTION_SPACE, then supply its parameters.
For example, move the cursor:
```json
{"action_type": "MOVE_TO", "parameters": {"x": 1319, "y": 65}}
```
Right-click at a specified position:
```json
{"action_type": "CLICK", "parameters": {"x": 960, "y": 540, "button": "right"}}
```
Press and release a key:
```json
{"action_type": "PRESS", "parameters": {"key": "q"}}
```
Press a key combination:
```json
{"action_type": "HOTKEY", "parameters": {"keys": ["ctrl", "q"]}}
```
Type text:
```json
{"action_type": "TYPING", "parameters": {"text": "hello world"}}
```
When the task is complete:
```json
{"action_type": "DONE", "parameters": {}}
```

REMEMBER:
Return ONLY the action_type and parameters requested above, in a JSON dictionary
wrapped in a json code block. Choose ONLY from ACTION_SPACE.
Do not return Python code or a JSON array.
You may predict multiple actions in one turn, with exactly one dictionary per code block.
Actions execute in their output order. DONE or FAIL must be the final action.
WAIT waits briefly; DONE ends the task; FAIL declares it infeasible. Try your best before FAIL.
The parser also accepts plain JSON dictionaries and bare or fenced WAIT/DONE/FAIL markers.

KEY_DOWN and MOUSE_DOWN keep the key or button held across actions and turns;
release them with KEY_UP and MOUSE_UP when appropriate.
The game is already prepared. Stay in its window.

League of Legends reminders:
1. Avoid staying in the fountain too long or spending prolonged periods without
   in-game actions, as this can trigger AFK detection.
   If an AFK warning appears, click the OK button in the center of the screen
   to return to the game controls.
2. You are playing with a locked camera centered on your champion.
3. When you learn a new ability or purchase a new item, you can read its description.
4. You use normal cast with targeting indicators, not quick cast.
\end{Verbatim}
\end{tcolorbox}

\section{Supplementary Results}
\label{app:matrix}
This appendix extends Section~\ref{sec:experiments} with per-task results, subgoal breakdowns, and latency statistics for the same nine configurations.

\subsection{Per-Task Results}
\begin{figure}[!htbp]
\centering
\includegraphics[width=\linewidth]{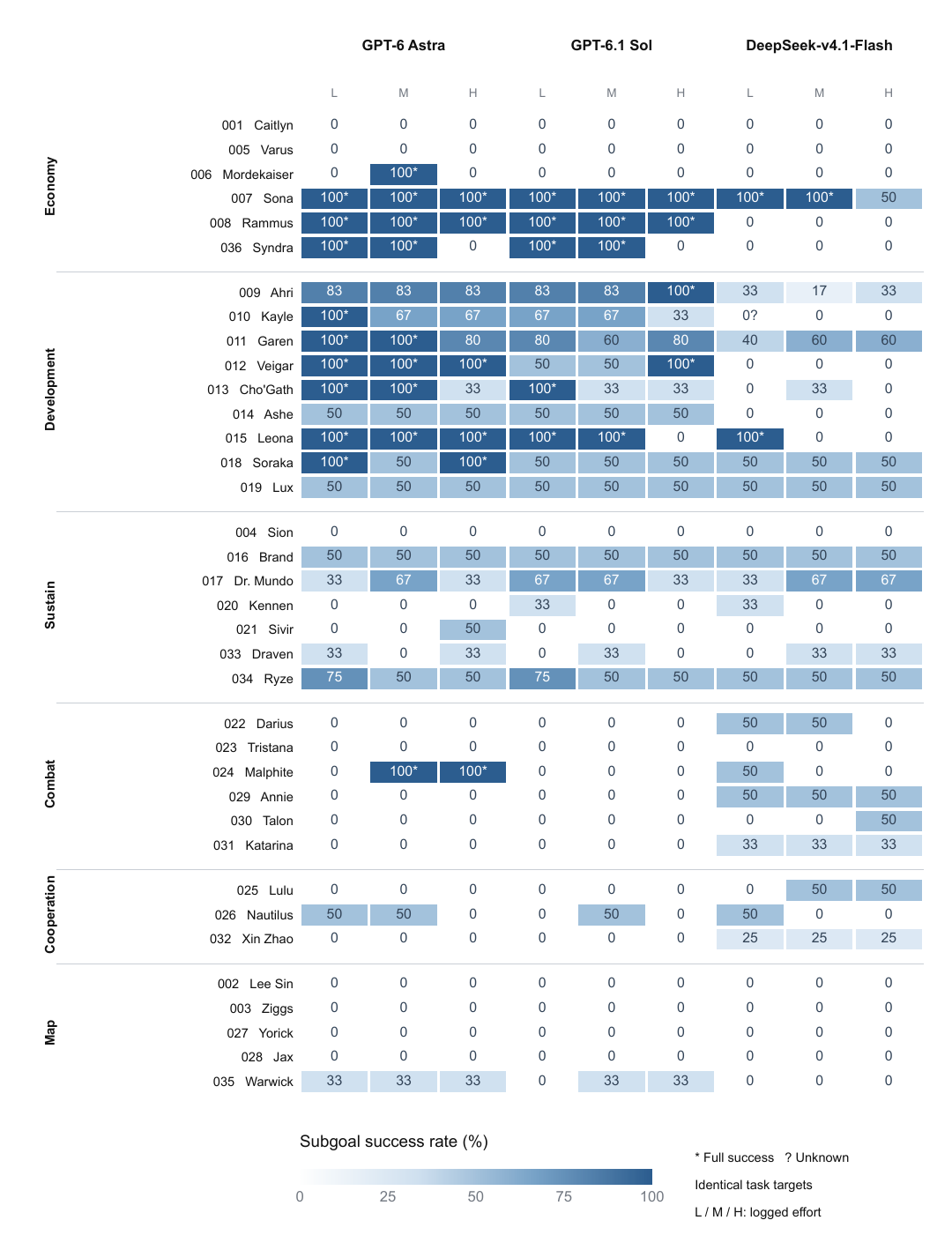}
\caption{\textbf{Complete 36-task diagnostic matrix.} Rows identify tasks and champions, grouped by primary demand; columns identify model and reasoning effort. Cells show subgoal success rate (\%, rounded); asterisks mark task success. The question mark identifies the setup failure with unavailable subgoal measurements. All configurations use identical task targets. A value of 100 does not by itself establish task success, which also requires valid timing and task context.}
\label{fig:matrix}
\end{figure}

Figure~\ref{fig:matrix} reports subgoal success rate and task outcomes for every task and configuration.
The union of passing task IDs is 006, 007, 008, 009, 010, 011, 012, 013, 015, 018, 024, and 036. These 12 tasks succeed in at least one configuration. The reported subgoal success rate gives each task equal weight. Pooling the checks instead gives 243/738 (32.9\%); each configuration contributes the same 82 checks.

\subsection{Subgoal Breakdown}
\label{app:subgoal-diagnostics}
This breakdown explains why satisfying some subgoals need not demonstrate progress toward completing the whole task (Section~\ref{sec:experiments}). It includes all 36 tasks and all 738 subgoal checks across the nine configurations; no task or check is excluded. We separate requirements for acquiring resources or completing events from requirements for avoiding violations or preserving a state.

\emph{Positive attainment} includes reaching a level, obtaining an item, or securing a kill. \emph{Avoidance/absence} includes staying below a limit or leaving an ability unlearned. \emph{State preservation} includes remaining alive or maintaining a required health or mana fraction. These categories describe the requirements checked, rather than attributing their outcomes to particular actions.

Table~\ref{tab:subgoal-semantics} reports 144/567 (25.4\%) achieved positive-attainment checks, 53/117 (45.3\%) avoidance/absence checks, and 46/54 (85.2\%) state-preservation checks. Thus, 99 of the 243 achieved checks (40.7\%) concern avoiding violations or preserving conditions. These are legitimate task requirements, but satisfying them alone cannot complete a task that also requires acquiring an item or securing an objective. This distinction helps interpret partial subgoal success alongside the low task success rate. The counts pool individual checks; the headline subgoal success rate instead gives each task equal weight.

\begin{table}[ht]
\centering\small
\caption{\textbf{Subgoal success rate by requirement type.} Cells give achieved/eligible checks (\%). Positive denotes positive attainment; avoidance includes upper bounds and exact-zero skill ranks; preservation covers alive and resource-fraction conditions. Rates pool checks within each type; Table~\ref{tab:main} instead averages subgoal success rate over tasks. All configurations use identical targets.}
\label{tab:subgoal-semantics}
\setlength{\tabcolsep}{4pt}
\begin{tabular*}{\linewidth}{@{\extracolsep{\fill}}llrrr@{}}
\toprule
Model & Effort & Positive attainment & Avoidance / absence & State preservation \\
\midrule
GPT-6 Astra & Low & 26/63 (41.3) & 4/13 (30.8) & 5/6 (83.3) \\
 & Medium & 25/63 (39.7) & 5/13 (38.5) & 5/6 (83.3) \\
 & High & 21/63 (33.3) & 5/13 (38.5) & 5/6 (83.3) \\
\midrule
GPT-6.1 Sol & Low & 21/63 (33.3) & 4/13 (30.8) & 5/6 (83.3) \\
 & Medium & 18/63 (28.6) & 4/13 (30.8) & 6/6 (100.0) \\
 & High & 17/63 (27.0) & 4/13 (30.8) & 4/6 (66.7) \\
\midrule
DeepSeek-v4.1-Flash & Low & 7/63 (11.1) & 9/13 (69.2) & 4/6 (66.7) \\
 & Medium & 5/63 (7.9) & 9/13 (69.2) & 6/6 (100.0) \\
 & High & 4/63 (6.3) & 9/13 (69.2) & 6/6 (100.0) \\
\bottomrule\end{tabular*}\end{table}

\begin{figure}[!htbp]
\centering
\includegraphics[width=\linewidth]{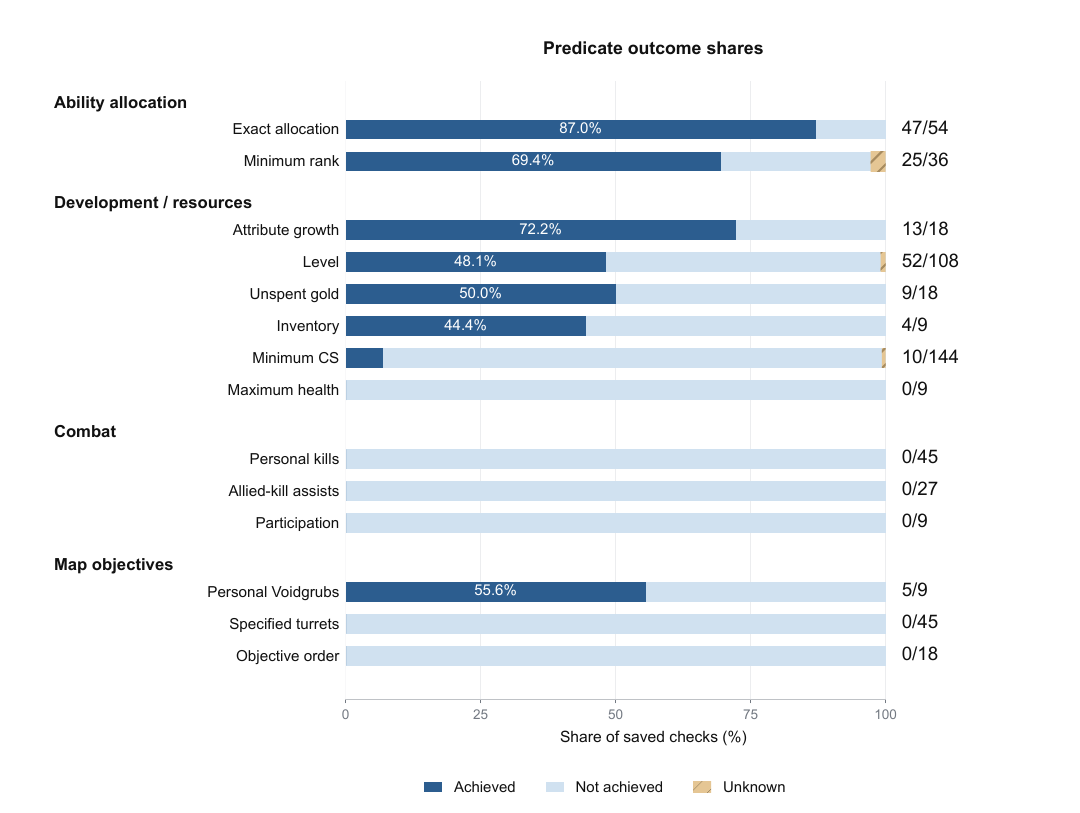}
\caption{\textbf{Subgoal outcomes by requirement.} Counts pool subgoal checks across configurations. ``Exact ability allocation'' combines positive allocations and intentionally unlearned abilities; unlike the minimum-rank row, it includes easy-to-satisfy zero constraints. Item checks require the specified inventory IDs.}
\label{fig:goaltypes}
\end{figure}

Figure~\ref{fig:goaltypes} provides the complementary breakdown by game property, allowing readers to identify which specific requirements are met.

\label{app:goals}
Together, the two breakdowns distinguish meeting individual requirements from completing their conjunction within a task. Neither changes the success criteria or removes achieved subgoals from the reported scores.

\subsection{Response Latency}
\label{app:timing}
This section supplements Figure~\ref{fig:temporal} with latency statistics pooled across the nine configurations.

Latency statistics use completed responses, including service, network, retry, and parsing time. Requests with no completed response are excluded. Responses and actions are counted separately because one response can contain several actions.

Across 17,221 completed requests, mean, median, and 95th-percentile latency are 6.536, 5.594, and 14.031 seconds, respectively. These measurements include network and service overhead; they do not measure internal reasoning time alone.

\Needspace{10\baselineskip}
\subsection{Response-Cap Analysis}
\label{app:response-cap}
Table~\ref{tab:response-cap} conditions on DeepSeek-v4.1-Flash episodes that end at the 100-response cap. Low and medium effort reach this cap after a median of 64.0\% and 63.7\% of their wall-time allowances, respectively. Short requests allow the agent to consume the response budget while game time remains; the subgoal breakdown shows whether those decisions also satisfy the task requirements.

\begin{table}[!htbp]
\centering\small
\setlength{\tabcolsep}{5pt}
\caption{\textbf{Progress when DeepSeek-v4.1-Flash reaches the response cap.} Counts cover only capped runs. Task success and zero-subgoal columns count runs. The last two columns count achieved/eligible checks, partitioned by the requirement types defined above.}
\label{tab:response-cap}
\begin{tabular}{@{}lrrrrr@{}}
\toprule
Effort & Capped runs & Task success & Zero subgoals & \shortstack{Positive\\attainment} & \shortstack{Avoidance /\\preservation} \\
\midrule
Low & 26 & 1 & 13 & 4/46 & 12/16 \\
Medium & 29 & 0 & 19 & 1/47 & 11/14 \\
High & 14 & 0 & 8 & 0/24 & 8/9 \\
\bottomrule
\end{tabular}
\end{table}

\paragraph{Many inputs, little task progress.}
In task~024, Malphite must reach level~5 with at most one death within nine minutes. DeepSeek (low) issues 104 actions over 100 responses: 101 right-click commands, one ability-learning hotkey, one cursor movement, and one key press. Observations after responses~50 and~100 both show level~1 near the allied base. At termination after 382.8 seconds of the 540-second budget, the final state has level~1 and zero deaths. Its 50\% subgoal success therefore reflects the death limit alone; the required leveling has not occurred. This case illustrates a mismatch between input volume and objective progress, without assuming that every right-click was ineffective.

The cap also has a direct consequence for tasks that require play through the deadline: six low-effort and five medium-effort capped episodes end before that required duration and receive an incomplete-duration outcome. These cases should be distinguished from tasks that permit early completion. Neither low partial progress nor this case study establishes the outcome of an uncapped continuation; all reported comparisons use the same time and response budgets.

\end{document}